\documentclass[sigconf]{acmart}
\usepackage{bbm}
\usepackage{tikz}
\usepackage{pgfplots}
\pgfplotsset{compat=1.18}
\usepgfplotslibrary{fillbetween}

\usepackage{array}
\usepackage{multirow}
\usepackage{multicol}

\usepackage{wrapfig}
\usepackage{tcolorbox}

\usepackage{etoolbox}
\BeforeBeginEnvironment{wrapfigure}{\setlength{\intextsep}{0pt}}
\usepackage{caption}
\AtBeginDocument{%
  }

\copyrightyear{2026}
\acmYear{2026}
\setcopyright{cc}
\setcctype{by}
\acmConference[KDD 2026] {Proceedings of the 32nd ACM SIGKDD Conference on Knowledge Discovery and Data Mining V.2}{August 9--13, 2026}{Jeju Island, Republic of Korea.}
\acmBooktitle{Proceedings of the 32nd ACM SIGKDD Conference on Knowledge Discovery and Data Mining V.2 (KDD 2026), August 9--13, 2026, Jeju Island, Republic of Korea}
\acmISBN{979-8-4007-2259-2/2026/08}
\acmDOI{10.1145/3770855.3817788}
\begin{document}

\title{Exploiting Similarities in A/B Testing with Off-Policy Estimation}

\author{Otmane Sakhi}
\orcid{0009-0004-6185-7574}
\email{o.sakhi@criteo.com}
\affiliation{%
  \institution{Criteo AI Lab}
  \city{Paris}
  \country{France}
}

\author{Alexandre Gilotte}
\orcid{0009-0003-7144-934X}
\email{a.gilotte@criteo.com}
\affiliation{%
  \institution{Criteo AI Lab}
  \city{Paris}
  \country{France}
}

\author{David Rohde}
\orcid{0000-0002-0661-6266}
\email{d.rohde@criteo.com}
\affiliation{%
  \institution{Criteo AI Lab}
  \city{Paris}
  \country{France}
}

\renewcommand{\shortauthors}{Otmane Sakhi,  Alexandre Gilotte, \& David Rohde}

\begin{abstract}
We study A/B testing, the standard protocol for measuring the performance gain of a new decision system relative to a baseline. Traditional A/B testing treats both systems as black boxes, ignoring potential similarities between them. In practice, however, new and baseline systems are rarely radically different and often share significant structure, which can be captured by their propensities to make similar decisions. We show that in such cases, the commonly used difference-in-means estimator, though unbiased, is statistically suboptimal. Leveraging off-policy estimation, we introduce a family of A/B testing estimators that exploit the propensities of the tested systems to achieve improved concentration properties. This family is flexible enough to be tailored to practical decision-making. The resulting estimators are simple, robust to propensities misspecification, substantially more accurate when the tested systems exhibit similarities, and gracefully fall back to the difference-in-means estimator when such similarities are absent. Our theoretical analysis and empirical studies confirm their efficiency and practicality.
\end{abstract}




\maketitle

\section{Introduction}

Online interactive systems are pervasive, with key applications in computational advertising \citep{bottou2013counterfactual}, search \citep{search}, and recommendation \citep{sakhi2020blob}, among many others. Their success relies on the ability to rapidly iterate on deployed systems, test new updates, and ensure that each iteration delivers measurable improvements \citep{m1,m2}. Central to this process are estimation protocols that quantify the expected benefits of proposed updates, enabling reliable decision-making and continuous system improvement.

A/B testing \citep{kohavi2012trustworthy} is widely regarded as the \emph{gold standard} protocol for measuring the improvement brought by a proposed change, due to its ease of use, weak assumptions, and reliability when best practices are followed \citep{kohavi2013online}. The protocol splits the user population into two groups: one exposed to the current system and the other to a new, potentially improved version. The improvement is then estimated as the difference between the empirical values of the two systems on their respective populations. Under standard assumptions, this difference-in-means estimator is unbiased, and its variance is primarily driven by the variability of the target signal. This variance plays a central role in decision-making: for a fixed sample size, increasing the sensitivity of an experiment requires reducing estimation variance. Several approaches based on regression adjustment with auxiliary data have been proposed to achieve this goal \citep{cuped,in_experiment}.

In parallel, a large body of work has studied the evaluation of system updates through the lens of policy evaluation and policy comparison \citep{bottou2013counterfactual}. In particular, off-policy estimation (OPE) \citep{sutton98reinforcement} has received significant attention, as it aims to evaluate new decision systems using only historical interactions, reducing the need for live experimentation. These methods typically rely on importance weighting and related counterfactual estimators \citep{horvitz1952generalization,bottou2013counterfactual}.

When the new system is sufficiently similar to the current one, off-policy methods can provide accurate estimates of the improvement brought by the update \citep{horvitz1952generalization,bottou2013counterfactual}. However, their reliability deteriorates as the compared systems become more different \citep{sakhi2024logarithmicsmoothingpessimisticoffpolicy}. In particular, support mismatch can induce bias \citep{sachdeva2020off}, while large or unstable importance weights can lead to high variance \citep{swaminathan2015counterfactual}. Recent work has attempted to mitigate these issues \citep{saito2022off,pol_conv}, but often by introducing additional assumptions that may limit applicability in real-world systems.

In this work, we take a different perspective. Rather than using importance weighting to replace A/B testing, we use it to improve A/B testing itself. This is particularly relevant in modern online applications, where decision systems are often stochastic by design, to enable exploration, personalization and improved diversity \citep{bottou2013counterfactual,agarwal2016multiworld, diversity}. In such settings, the systems' propensities are available, or can often be estimated, and provide useful information about how similarly the two policies behave. We first show that, when the tested policies are similar, importance weighting can yield more accurate improvement estimates than the standard difference-in-means estimator. This gain arises because the classical A/B test estimator treats the two systems as black boxes and ignores information about their propensities to take similar decisions. Motivated by this observation, we develop improved A/B testing estimators that exploit similarities between the tested policies, leading to the following contributions.

% \paragraph{\textbf{Contributions.}} We propose an intuitive construction of A/B testing estimators based on importance weighting. This construction yields a family of estimators that provably improve upon the mean squared error of the difference-in-means estimator whenever the compared policies exhibit overlap, while naturally reducing to difference-in-means when the policies are dissimilar. Our approach requires access only to policy propensities and can be made robust to propensity misspecification when these are estimated from data. The proposed estimators rely on no additional assumptions, apply to realistic settings, and can directly replace standard A/B testing whenever propensities are available. We study this family theoretically, identifying estimators with substantially improved concentration properties under policy overlap, and validate their effectiveness through experiments in practical scenarios.

\paragraph{\textbf{Contributions.}}
We propose an intuitive construction of A/B testing estimators based on importance weighting. This construction yields a flexible family of estimators that exploit overlap between the tested policies while naturally reducing to the standard difference-in-means estimator when no useful overlap is present. Our approach requires only access to policy propensities, introduces no changes to the standard A/B testing protocol, and can be made robust to propensity misspecification when propensities are estimated from data. We provide a theoretical analysis of this family, identify estimators with improved concentration properties under policy overlap, and derive misspecification-aware variants for practical deployment. Empirically, we show that the proposed estimators substantially reduce mean squared error in realistic scenarios where the tested policies share structure.

\paragraph{\textbf{Related Work.}}
Off-policy estimation (OPE) has a long history in evaluating decision rules from logged interactions. Early approaches relied on the Horvitz--Thompson estimator \citep{horvitz1952generalization}, and more generally inverse propensity scoring, to estimate the average reward of new decision rules without deploying them on live traffic \citep{beygelzimer2011contextual,bottou2013counterfactual,swaminathan2015counterfactual}. These methods were influential as one of the first attempts to quantify the business impact of a new policy from historical data alone, but they were mostly developed in simplified settings.

Most OPE methods for online decision systems are studied in contextual bandit settings with small action spaces and sufficient overlap between the logging and target policies \citep{joachims2018deep}. Even in these favorable regimes, importance-weighted estimators can suffer from large variance and sensitivity to poor support overlap, motivating extensive work on weight control and stabilization \citep{aouali23a,sakhi2024logarithmicsmoothingpessimisticoffpolicy}. Despite these advances, OPE remains difficult to apply reliably in realistic environments \citep{gilotte2018offline}, particularly in more general sequential settings \citep{kallusdouble2020}. As a result, it cannot generally replace the gold-standard A/B testing protocol. Our work follows a complementary direction: rather than using OPE as a substitute for A/B testing, we use OPE ideas to improve the estimator used within a standard A/B test by exploiting similarities between the tested policies.

Several works have also explored the use of off-policy techniques within online experimentation. \citet{agarwal2016multiworld} leverage IPS to reuse data collected under past policies, while \citet{pmlr-v162-wan22b} propose safe and efficient data-collection strategies for policy evaluation and comparison. These approaches are effective, but they typically rely on contextual bandit assumptions and modify the data-collection protocol. We keep the standard A/B testing protocol unchanged: both policies are deployed on separate user populations, and the estimator is modified only at the analysis stage. This allows us to exploit policy overlap while preserving the practical simplicity and reliability of standard A/B testing, and to account for propensity misspecification when propensities are estimated.

Recent work on online decision systems has addressed \emph{Markovian interference} \citep{farias_neurips}, where the Stable Unit Treatment Value Assumption (SUTVA) fails due to temporal or cross-user dependencies. These methods model interference using Markov decision processes \citep{mdp} and propose new experimental designs or corrected estimators \citep{farias_neurips,farias_dq,shi2021time,chen2024experimenting}. Our setting is orthogonal. We assume independent users, and therefore retain SUTVA at the user level, but allow each user's outcome to depend on a sequence of policy-driven interactions. Under this assumption, we improve A/B testing estimators by exploiting similarity between the tested policies, without modifying the experimental design.

Finally, our work is closely related to variance reduction techniques in classical A/B testing \citep{cuped,in_experiment,in_out}. These methods use auxiliary signals, such as pre-experiment or in-experiment covariates, to construct control variates and reduce outcome variance. While effective, they treat the tested systems as black boxes and do not use the fact that two decision policies may take similar actions. Our approach is complementary: it reduces variance through policy overlap, and can be combined with regression adjustment or outcome modeling, yielding estimators analogous to doubly robust methods \citep{dudik14doubly} with further variance reduction.

\section{Setting} 
We model the online decision problem, and specifically \emph{the dynamics of each user} with an extension of the non-Markov Decision Process (NMDP) framework \cite{mdp, kallusdouble2020}. Let $\mathcal{S} \subset \mathbb{R}^d$ be the \emph{state} space which is a compact subset of $\mathbb{R}^d$, and let $\mathcal{A}$ be our \emph{action set}. This action set can be either finite or continuous. The agent, representing our decision system is guided by a \emph{stochastic} and \emph{stationary} policy $\pi \in \Pi$ within a policy space $\Pi$. Given a state $s \in \mathcal{S}$, $\pi(\cdot | s)$ is a probability distribution over the action set $\mathcal{A}$; $\pi(a|s)$ is the probability that the agent selects action $a$ in state $s$. The decision system interacts with users \emph{independently}. For a user, the interaction begins with an initial state $s_0 \in \mathcal{S}$ coming from an unknown distribution $d_0(\cdot)$. At each time step $t$, the agent observes the user’s current state $s_t$, selects an action $a_t \sim \pi(\cdot | s_t)$, and receives a stochastic reward $r_t \in [0, 1]$, sampled from an unknown distribution $p(\cdot | s^u_t, a^u_t)$ of expectation $r(a_t, s_t)$. Then, the interaction at time $t$ transitions the user to the next state $s_{t+1}$ following an unknown transition kernel $\mathcal{K}(\cdot|[s_0,...,s_t], a_t)$\footnote{This framework accounts for non-Markovian, long term dependencies between states and outcomes.}. These rounds of interactions continue till we reach a terminal state $\bar{s} \in \mathcal{S}$ at time $T$, defining a complete user trajectory $\tau = \{(s_t, a_t, r_t)\}_{t \in [T]}$, guided by the agent's policy $\pi$. We assume that actions taken by the decision system $\pi$ influence only the next state of the same user, $s_{t+1}$, not affecting other users. For notational convenience, we denote by $\nu(\pi)$ the trajectory distribution induced by $\pi$ in the extended MDP.

 The performance of a policy $\pi$ is determined by its \emph{value} defined as the expectation of the cumulative reward computed over trajectories coming from the distribution $\nu(\pi)$ induced by $\pi$. The \emph{value} writes:
\begin{align}\label{eq:value}
    V(\pi)= \mathbb{E}_{\tau \sim  \nu(\pi)} \left[\sum_{t = 0}^T r(a_t, s_t) \right].
\end{align}

\paragraph{\textbf{A/B Testing.}} We are interested in comparing two policies $\pi_A$ and $\pi_B$, and measure whether the target policy $\pi_A$ is \emph{significantly} better than our base policy $\pi_B$. We define the true improvement brought by $\pi_A$, over $\pi_B$ as the difference between their two values, such as:
\begin{align}\label{eq:improvement}
    \mathcal{I}(\pi_A, \pi_B) = V(\pi_A) - V(\pi_B).
\end{align}
The objective of an A/B test is to build an estimate of this improvement and leverage it to \emph{statistically} test whether or not, the target policy $\pi_A$ is \emph{significantly} better than $\pi_B$. We build our improvement estimate using collected, \emph{independent} trajectories of interactions $\mathcal{D}_A$ (respectively $\mathcal{D}_B$) of the policy $\pi_A$ (respectively $\pi_B$) of the following form:
\begin{align*}
\mathcal{D}_A = \{\{ s^i_t, a^i_t, r^i_t \}_{t \in [T_i]}\}_{i \in u_A}\,, \quad \mathcal{D}_B = \{\{ s^j_t, a^j_t, r^j_t \}_{t \in [T_j]}\}_{j \in u_B}\,.
\end{align*}

Where $u_A$ and $u_B$ are disjoint subsets of users allocated to $\pi_A$ or $\pi_B$ respectively. $n_A$ is the number of users in $u_A$ and $n_B$ is the number of users in $u_B$. The classical improvement estimator is defined as the difference of the empirical estimates of the values. Particularly, we first leverage the collected data $\mathcal{D}_A$ by policy $\pi_A$ (respectively $\mathcal{D}_B$ by policy $\pi_B$), to compute their estimated values:
\begin{align*}
    \hat{V}(\pi_A) = \frac{1}{n_A} \sum_{i \in u_A} \sum_{t = 1}^{T_i} r^i_t\,, \quad \hat{V}(\pi_B) = \frac{1}{n_B} \sum_{j \in u_B} \sum_{t = 1}^{T_j} r^j_t\,.
\end{align*}

These estimates are naturally used to define the improvement estimator as their respective difference:
\begin{align}\label{eq:improvement_est}
    \hat{\mathcal{I}} = \hat{V}(\pi_A) - \hat{V}(\pi_B)\,.
\end{align}

% The difference-in-means estimator is \emph{unbiased} and its variance (or its empirical counterpart) will determine the statistical significance of the improvement \citep{alpha_test}. To improve our A/B testing approaches, the general goal is to find estimators with better concentration properties, generally quantified by a better mean squared error, which in the unbiased case, estimators with smaller variance. This will be explored next.
The difference-in-means estimator is \emph{unbiased}, and its variance directly determines the statistical significance of the observed improvement \citep{alpha_test}. Consequently, improving A/B testing methodology amounts to constructing estimators with improved concentration properties, typically measured through a lower mean squared error. In the unbiased setting, this reduces to designing estimators with smaller variance. We investigate this in the following section.
% \begin{align}\label{eq:test}
%     H_0: \mathcal{I}(\pi_A, \pi_B) \le 0 \quad \text{  v.s.  } \quad H_1: \mathcal{I}(\pi_A, \pi_B) > 0.
% \end{align}
% We denote by $\sigma^2(\hat{\mathcal{I}})$ the variance of the improvement estimator and suppose that $n_A = n_B = n$. Using  the Central Limit Theorem and Slutsky Lemma, we can be confident of the improvement with an (asymptotic) $\alpha$ level test \citep{alpha_test} when:
% \begin{align*}
%     \hat{\sigma}(\hat{\mathcal{I}})^{-1} \sqrt{n} \, \hat{\mathcal{I}} \ge z_{1 - \alpha}\,,
% \end{align*}

% where $z_{1 - \alpha}$ is the upper $\alpha$-th quintile of the standard normal distribution. For any $\Delta > 0$, the (asymptotic) power of our test under the local alternative $H_{1,n}: \mathcal{I}(\pi_A, \pi_B) = \Delta/\sqrt{n}$ is consequently $1 - \Phi(z_{1-\alpha} - \sigma(\hat{\mathcal{I}})\Delta)$, where $\Phi(\cdot)$ is the cumulative distribution function of a normal Gaussian. This means that under the same budget, and since all terms are constant, the power of our test is fully determined by the variance of the improvement estimator.  

\section{Motivating Example}

% \subsection{The Motivation}

We study the variance of the difference-in-means estimator defined in Equation \eqref{eq:improvement_est}. To simplify our argument, let us suppose that $n_A = n_B = n$ for the rest of the analysis. By exploiting the independence assumption, one can write down the estimator's variance as:
\begin{align*}
    n \sigma^2(\mathcal{\hat{I}}) =\text{var}_{\nu(\pi_A)}\left[ \sum_{t = 1}^T r_t \right] + \text{var}_{\nu(\pi_B)}\left[ \sum_{t = 1}^T r_t \right],
\end{align*}
with $\text{var}_{\nu(\pi_A)}$ (resp. $\text{var}_{\nu(\pi_B)}$) the variance of the cumulative reward under trajectories generated by $\pi_A$ (resp. $\pi_B$). The variance of this estimator will be equal to $0$ if and only if the reward signal is constant \emph{and} the transition kernel is deterministic. This case never happens in practice as it means that all policies are equivalent for the problem in hand. Therefore, the classical difference estimator will always suffer non negligible variance, primarily driven by the variance of the cumulative reward.

\paragraph{\textbf{A/A Test and Overlapping Policies.}} Imagine that we compare identical policies ($\pi_A = \pi_B$). The true improvement $\mathcal{I}(\pi_A, \pi_B)$ is equal to $0$ in this case, the classical difference-in-means estimator $\hat{\mathcal{I}}$ is centered around $0$ and suffers a variance of:
 \begin{align*}
    n \sigma^2(\mathcal{\hat{I}}) = 2 \text{ var}_{\nu(\pi_A)}\left[ \sum_{t = 1}^T r_t \right].
\end{align*}
The variance of the estimator is still of the same order of magnitude as before even if $\pi_A$ and $\pi_B$ are identical. This is a sub-optimal behavior that can be improved. Importance weighting approaches \cite{horvitz1952generalization, bottou2013counterfactual} were demonstrated to produce low variance estimators when the tested policies overlap and can be leveraged in these scenarios. We begin by defining the following quantities as they will be of interest in the rest of this work. 
% \begin{definition}(Trajectory propensities and importance weights.)
Let $t \in \mathbb{N}^*$, and let $\tau_t$ be a trajectory composed of states and performed actions up to time step $t$: $\tau_t = \{s_1, a_1, \dots, s_t, a_t \}$. We define the propensity of a trajectory $\tau_t$ under a policy $\pi$ by the following:
\begin{align}
    \pi(\tau_t) = \prod_{l = 1}^t \pi(a_l|s_l).
\end{align}
We define $w_A(\tau_t)$, the importance weight of a trajectory $\tau_t$ collected under $\pi_A$ ($\pi_A(\tau_t) > 0$), and evaluated with $\pi_B$ and respectively, we define $w_B(\tau_t)$, the importance weight of a trajectory $\tau^t$ collected under $\pi_B$ ($\pi_B(\tau^t) > 0$), and evaluated with $\pi_A$:
\begin{align}
    w_A(\tau_t) = \frac{\pi_B(\tau_t)}{\pi_A(\tau_t)} \in \mathbb{R}^+\,, \quad w_B(\tau_t) = \frac{\pi_A(\tau_t)}{\pi_B(\tau_t)} \in \mathbb{R}^+\,.
\end{align}
Finally, when $\tau_t$ has a non null propensities under both policies, we have: $w_B(\tau_t) w_A(\tau_t)= 1$. With these definitions, let us write down an Inverse Propensity Scoring estimator (IPS) \cite{horvitz1952generalization} of the improvement using the collected interactions $\mathcal{D}_B$:
\begin{align}\label{eq:naive_ips}
    \hat{\mathcal{I}}_{\textsc{IPS}, \mathcal{D}_B} = \frac{1}{n} \sum_{j \in u_B } \sum_{t = 1}^{T_j} (w_A(\tau^j_t) - 1) r^j_t\,.
\end{align}
If $\pi_A$ and $\pi_B$ have a common support, it is straightforward to prove that the IPS estimator is unbiased. We observe that we can achieve substantial variance reduction compared to the difference estimator of Equation \eqref{eq:improvement_est}, when $\pi_A$ and $\pi_B$ are similar, translating to $\forall \tau,\, w_A(\tau) \approx 1$. Precisely, in the special case of comparing identical policies, the obtained estimator as well as its variance reduce exactly to $0$. This proves that the sub-optimality previously identified of the difference estimator can be corrected, and highlights the advantage of using importance weighting in an A/B testing scenario. However, this estimator, as advantageous as it can be, will suffer from substantial bias and variance in practical settings. 

\textbf{Bias.} The common support assumption is restrictive and does not hold in practice \cite{sachdeva2020off}. The two systems may deliver similar actions on some states, and in other states differ. Our goal is to develop \emph{unbiased} estimators that leverage this inherent structure without relying on additional assumptions. 

\textbf{Variance.} Intuitively, the variance of this estimator is small in trajectories where the two policies are similar, but the global variance is dominated by trajectories where the policies differ \cite{sakhi2024logarithmicsmoothingpessimisticoffpolicy}. This makes the proposed IPS estimator less appealing than the difference-in-means in practical scenarios. Our goal is to develop an estimator that exploits the similarity structure to reduce variance, while matching the variance of the difference estimator at worst.

\section{Off-Policy Augmented A/B Testing}

\subsection{Construction of our family of estimators}

Direct application of importance weighting produces estimators with substantial bias and variance in practice \cite{sachdeva2020off}. To address the variance problem, numerous importance weight transforms were developed that enable effective bias-variance trade-offs, defining the large family of regularized IPS estimators \cite{bottou2013counterfactual,su2020doubly,metelli2021subgaussian,aouali23a, sakhi2024logarithmicsmoothingpessimisticoffpolicy}. While these transforms successfully reduce variance, they introduce additional bias that must be accounted for \cite{gilotte2018offline, sakhi2023pac}. For instance, when using data $\mathcal{D}_B$ collected under policy $\pi_B$ to construct the regularized IPS estimator of the improvement, the bias suffered can be expressed as an expectation under trajectories from policy $\pi_A$. Fortunately, in A/B testing, we have access to $\mathcal{D}_A$, trajectories collected under $\pi_A$, allowing us to estimate and correct this bias.

Formally, let $f : \mathbb{R}^+ \rightarrow \mathbb{R}$ be the importance weights transform characterizing our novel off-policy estimator. We choose our function to be bounded, respecting $\lim_{x \rightarrow +\infty} f(x) \in \mathbb{R}.$ We also adopt the convention $f(1/0^+) = f(+\infty)$. With this function $f$, we introduce our estimator:
\begin{align*}
    \hat{\mathcal{I}}_{f} &= \underbrace{\frac{1}{n_B} \sum_{j \in u_B}\sum_{t = 1}^{T_j} f(w_A(\tau^j_t)) r^j_t}_{\text{$f$-Regularized IPS}} \\ &+ \underbrace{\frac{1}{n_A} \sum_{i \in u_A} \sum_{t = 1}^{T_i} \left(1 - w_B(\tau^i_t)\left[ 1 + f(w_A(\tau^i_t))\right] \right) r^i_t}_{\text{Bias Correction}}\,,
\end{align*}
Our estimator $\hat{\mathcal{I}}_{f}$ follows the previously outlined construction. Its first term corresponds to the $f$-regularized IPS estimator of the improvement using the interactions $\mathcal{D}_B$, while its second term corresponds to the bias correction computed using $\mathcal{D}_A$. 

\paragraph{\textbf{Remark.}} We can consider applying the same approach in reverse: using $\mathcal{D}_A$ to compute an $f'$-regularized IPS estimator and correcting its bias using $\mathcal{D}_B$. Furthermore, we could even think of defining a convex combination of the two estimators with $\lambda \in [0, 1]$. Appendix~\ref{app:conv_same_family} proves this approach is equivalent to using a function $g$ with our estimator, demonstrating that our current parametrization is already efficient in leveraging both sets of interactions.

\subsection{Properties of our family of estimators}

\paragraph{\textbf{Unbiased estimator.}} Our estimator properties are determined by the choice of $f$. The following proposition characterizes the condition on $f$ that yields an unbiased estimator.

\begin{proposition}[Unbiased family of estimators]{} \label{unbiased_family}
Let $f: \mathbb{R}^+ \rightarrow \mathbb{R}$ be a bounded function. $\hat{\mathcal{I}}_{f}$ is an unbiased estimator of $\mathcal{I}(\pi_A, \pi_B)$ when $f$ respects the additional condition:
\begin{align}\label{eq:cond_1}
    f(0) = -1\,. \tag{$C_1$}
\end{align}
\end{proposition}

\begin{proof}
Let $f$ be bounded. Taking expectations under trajectories from
$\pi_A$ and $\pi_B$, we have
\begin{align*}
\mathbb{E}\!\left[\hat{\mathcal I}_f\right]
&=
\mathbb{E}_{\nu(\pi_B)}
\!\left[
\sum_{t=1}^T f(w_A(\tau_t)) r_t
\right] \\
&\quad+
\mathbb{E}_{\nu(\pi_A)}
\!\left[
\sum_{t=1}^T
\left\{
1 - w_B(\tau_t)\bigl(1+f(w_A(\tau_t))\bigr)
\right\} r_t
\right] \\
&=
\mathbb{E}_{\nu(\pi_B)}
\!\left[
\sum_{t=1}^T
\left(
f(w_A(\tau_t))
-
\mathbbm{1}[\pi_A(\tau_t)>0]
\bigl(1+f(w_A(\tau_t))\bigr)
\right) r_t
\right] \\
&\quad+ V(\pi_A) \\
&=
\mathbb{E}_{\nu(\pi_B)}
\!\left[
\sum_{t=1}^T
\left(
\mathbbm{1}[\pi_A(\tau_t)=0] f(0)
-
\mathbbm{1}[\pi_A(\tau_t)>0]
\right) r_t
\right] + V(\pi_A)\\
&=
\mathcal I(\pi_A,\pi_B)
+
\mathbb{E}_{\nu(\pi_B)}
\!\left[
\sum_{t=1}^T
\mathbbm{1}[\pi_A(\tau_t)=0]\bigl(f(0)+1\bigr) r_t
\right].
\end{align*}
Therefore, imposing $f(0)=-1$, which is precisely
\eqref{eq:cond_1}, makes
$\hat{\mathcal I}_f$ an unbiased estimator of
$\mathcal I(\pi_A,\pi_B)$.
\end{proof}

% \begin{proof}
% Let $f$ be a bounded function. We take the expectation under trajectories of $\pi_A$ and $\pi_B$ and develop:
% \begin{align*}
%  \mathbb{E}_{\nu(\pi_A), \nu(\pi_B)} \left[ \hat{\mathcal{I}}_{f}\right] &= \mathbb{E}_{\nu(\pi_B)} \left[\sum_{t = 1}^{T} f(w_A(\tau_t)) r_t \right] \\ 
% &+ \mathbb{E}_{\nu(\pi_A)} \left[ \sum_{t = 1}^{T} \left(1 - w_B(\tau_t)\left[ 1 + f(w_A(\tau_t))\right] \right) r_t\right] \\
% &= \mathbb{E}_{\nu(\pi_B)} \left[\sum_{t = 1}^{T} f(w_A(\tau_t)) r_t \right] \\ 
% &- \mathbb{E}_{\nu(\pi_A)} \left[ \sum_{t = 1}^{T} w_B(\tau_t)\left[ 1 + f(w_A(\tau_t))\right] r_t\right] + V(\pi_A) \\
% &= \mathbb{E}_{\nu(\pi_B)} \left[\sum_{t = 1}^{T} f(w_A(\tau_t)) r_t \right] \\ &- \mathbb{E}_{\nu(\pi_B)} \left[ \sum_{t = 1}^{T} \mathbbm{1}[\pi_A(\tau_t) > 0]\left[ 1 + f(w_A(\tau_t))\right] r_t\right] + V(\pi_A) \\
% &= \mathbb{E}_{\nu(\pi_B)} \left[\sum_{t = 1}^{T} \mathbbm{1}[\pi_A(\tau_t) = 0] f(w_A(\tau_t)) r_t \right] - \mathbb{E}_{\nu(\pi_B)} \left[ \sum_{t = 1}^{T} \mathbbm{1}[\pi_A(\tau_t) > 0] r_t\right] + V(\pi_A) \\
%  &= \mathbb{E}_{\nu(\pi_B)} \left[\sum_{t = 1}^{T} \mathbbm{1}[\pi_A(\tau_t) = 0] (f(0) + 1) r_t \right] + \mathcal{I}(\pi_A, \pi_B).
% \end{align*}
% Meaning that by imposing $f(0) = -1$, which is exactly \eqref{eq:cond_1}, we obtain an unbiased estimator. This concludes the proof.
% \end{proof}

The condition stated in the previous proposition is simple and will help us design unbiased improvement estimators. For instance, let $h_{-}$ be the constant function $ h_{-}: x \rightarrow = -1$. $h_{-}$  respects condition \eqref{eq:cond_1} and setting $f = h_{-}$ results in the unbiased, difference-in-means estimator, \emph{i.e.} $\hat{\mathcal{I}}_{h_{-}} = \hat{\mathcal{I}}.$ This means that better choices of $f$ will lead to better behaved estimators. As we are seeking estimators that improve on the difference estimator, a question that arises is how do we choose $f$ to obtain an estimator with reduced variance.

\paragraph{\textbf{Lower variance.}} Let us first write down the variance of our general estimator. For any $f$, and by exploiting the independence assumption, we can decompose the variance of $\hat{\mathcal{I}}_{f}$ to:
\begin{align*}
    V_f = &\frac{1}{n_B} \text{var}_{\nu(\pi_B)}\left[\sum_{t = 1}^{T} f(w_A(\tau_t)) r_t \right] \\&+ \frac{1}{n_A} \text{var}_{\nu(\pi_A)}\left[ \sum_{t = 1}^{T} \left(1 - w_B(\tau_t)\left[ 1 + f(w_A(\tau_t))\right] \right) r_t \right].
\end{align*}
The first desired behavior is for the estimator to suffer no variance when the tested policies are identical. In this case, all importance weights are equal to $1$ and a null variance is obtained when:
\begin{align}\label{eq:cond_2}
    f(1) = 0\,. \tag{$C_2$}
\end{align}
Condition \eqref{eq:cond_2} ensures that the variance of our estimator improves on the difference estimator in the case of identical policies. For the general case, studying the variance of $\hat{\mathcal{I}}_{f}$ is a challenging task, and results, in conditions on $f$ that rely on intractable, problem-dependent quantities. This discussion is developed in Appendix~\ref{app:variance_surrogate}. As we seek simple conditions on $f$ that can achieve lower variance, we take the route of replacing this variance with a surrogate statistic $S_f$ that we define next, and that encodes the second moment behavior of the estimator.

\begin{definition}[Variance Surrogate]\label{def:surrogate}
Let $f: \mathbb{R}^+ \rightarrow \mathbb{R}$ be a bounded function, We define $S_f$ as:
\begin{align*}
    S_f = &\frac{1}{n_B} \mathbb{E}_{\nu(\pi_B)}\left[\sum_{t = 1}^T \left(f(w_A(\tau_t)) r_t \right)^2\right] \\&+  \frac{1}{n_A} \mathbb{E}_{\nu(\pi_A)}\left[\sum_{t = 1}^T \left(1 - w_B(\tau_t)\left[ 1 + f(w_A(\tau_t))\right] \right)^2 r_t^2\right].
\end{align*}
\end{definition}
The variance surrogate $S_f$ is constructed as the sum over the trajectories, of the individual second moment of each independent term of our estimator (of $\pi_A$ and $\pi_B$). $S_f$ is the second moment of our estimator, when the covariances within trajectories can be neglected \cite{lattimore19bandit}. In this case, this surrogate defines a valid, pessimistic upper bound on the variance. This upper bound can be tight and matches the variance in settings with sparse rewards, similar tested policies and small improvements. This is further detailed in Appendix~\ref{app:variance_surrogate} and demonstrates that our surrogate statistic will replicate the behavior of the variance in practical scenarios. 

We compare the variance surrogate of our estimators. Recall that the difference estimator is recovered by setting $f = h_{-}$. The next proposition identifies a condition on $f$ that reduces the variance surrogate of the difference estimator. 

\begin{figure}
    \centering

\resizebox{0.4\textwidth}{!}{%
    \begin{tikzpicture}
\begin{axis}[
    xlabel={$x$},
    ylabel={$f(x)$},
    xmin=-0.5, xmax=5,
    ymin=-1.3, ymax=1.2,
    samples=500,
    grid=major,
    legend pos=outer north east,
    ylabel style={yshift=-0.5cm},
    y label style={at={(axis description cs:-0.1,0.5)},rotate=270},
    yticklabels={,{$-1$},,,,{$1$}}
]

% \addplot[color=red, thick, name path=func3, domain=0:10] {-2 + 2 / (1 + exp(-x))} node[pos=0.2, sloped, above] {$g_1(x)$};
\addplot[color=blue, thick, name path=func3, domain=0:10] {min(x-1, 1)} node[pos=0.23, sloped, below] {$h_1(x)$};
% \addplot[color=orange, thick, name path=func3, domain=0:10] {min(pow(x,2)-1, 1)} node[pos=0.2, sloped, above] {};
\addplot[color=red, thick, name path=func3, domain=0:10] {(x-1)/(x + 1)} node[pos=0.3, sloped, below] {$f^\star_1(x)$};
% \addplot[color=red, thick, name path=func3, domain=0:10] {max(-1, min(0.5*(x-1), 1))} node[pos=0.2, sloped, above] {};

% Constant function h_-(x) = -1
\addplot[color=black, thick, dashed, name path=func1, domain=0:5] {-1} node[pos=.5, sloped, below] {$h_{-}(x)$};

% Piecewise linear function h_+(x) = min(2x-1, 1)
\addplot[color=black, thick, dashed, name path=func2, samples=500, domain=0:5] {min(2*x-1, 1)} node[pos=0.2, sloped, above] {$h_{+}(x)$};

\addplot[color=black, mark=*, mark size=1.7] coordinates {(0,-1)};
\addplot[color=black, mark=*, mark size=1.7] coordinates {(1,0)};

% Shaded area between functions
\addplot[color=gray, fill=gray!20, opacity=0.5, domain=0:10] fill between[of=func1 and func2];
\end{axis}
\end{tikzpicture}
}
    \caption{Surrogate reduction region defined by $h_{-}$ and $h_+: x \rightarrow \min(2x - 1, 1)$.}
    \label{fig:cond_1}
\end{figure}

% \Dave{Can you add a legend to Fig 1?}

\begin{proposition}[Variance surrogate reduction]{} \label{variance_reduction}
Let \eqref{eq:cond_3} be the condition on $f$ defined as:
\begin{align}\label{eq:cond_3}
        \forall x\in \mathbb{R}^+,\quad -1 \le f(x) \le \min\left(2x - 1, 1 \right)\,. \tag{$C_3$}
\end{align}
The following holds: $f \text{ respects } (C_3) \implies S_f \le S_{h_{-}}\,.$
\end{proposition}

The proof of this result is straightforward by controlling the two terms of the variance surrogate. The new condition \eqref{eq:cond_3} is sufficient to reduce the variance surrogate and will help us design, combined with the previous conditions defined (\eqref{eq:cond_1} and \eqref{eq:cond_2}), an estimator with better properties than the difference estimator. These conditions are not restrictive and result in an infinite pool of functions that recover an improved estimator, as shown in Figure~\ref{fig:cond_1}. We study one such function $f$, defined:
\begin{align}\label{eq:mae}
    \forall x \in \mathbb{R}^+, \quad h_1(x) = \min(x - 1, 1)\,.
\end{align}
$h_1$ represents a valid choice, respecting all previous conditions and uses a clipping function, reminiscent of the clipped IPS estimator \cite{bottou2013counterfactual}, fixing the clipping constant $M = 1$. $h_1$ clips the importance weights take $1$ as we are interested by the improvement and not the value. We write down the estimator:
\begin{align*}
    \hat{\mathcal{I}}_{h_{1}} = &\underbrace{\frac{1}{n_B} \sum_{j \in u_B} \sum_{t = 1}^{T_j} \min\left(w_A(\tau^j_t) - 1, 1\right) r^j_t}_{\text{Clipped IPS with $h_1$}}  \\ &+ 
    \underbrace{\frac{1}{n_A} \sum_{i \in u_A} \sum_{t = 1}^{T_i} \mathbb{I}\left[w_B(\tau^i_t) < \frac{1}{2}\right]\left(1 - 2w_B(\tau^i_t)\right) r^i_t}_{\text{$h_1$ Bias Correction}}\,.
\end{align*}

This estimator should improve on the variance of the difference estimator as its variance will be negligible in states where the policies behave similarly (i.e. $w_A(\tau^t) \approx w_B(\tau^t) \approx 1$) and comparable to the difference estimator when they play different actions (i.e. $w_A(\tau^t) \approx w_B(\tau^t) \approx 0$). Note that if the tested policies are close but not identical, for instance, $1/2 <w_A(\tau^t)< 2$ we will also have $1/2 < w_B(\tau^t)< 2$ and our bias-corrected clipping estimator will only use data coming from $\mathcal{D}_B$, which is a suboptimal behavior. Other functions can be used to define our estimator \cite{aouali23a, sakhi2024logarithmicsmoothingpessimisticoffpolicy}, as long as the previously defined conditions are respected.

\subsection{Surrogate-Optimal Improvement Estimator}

\begin{figure}
    \centering
    \resizebox{0.4\textwidth}{!}{%
    \begin{tikzpicture}
\begin{axis}[
    xlabel={$x$},
    ylabel={$f(x)$},
    xmin=-0.5, xmax=10,
    ymin=-1.5, ymax=2.5,
    samples=500,
    grid=major,
    legend pos=outer north east,
    ylabel style={yshift=-0.5cm},
    y label style={at={(axis description cs:-0.1,0.5)},rotate=270},
]

\addplot[color=black, thick, dashed, name path=func2, samples=500, domain=0:20] {min(2*x-1, 1)} node[pos=0.18, sloped, above] {$h_{+}(x)$};
\addplot[color=blue, thick, name path=func3, domain=0:20] {(x - 1)/(x + 1)} node[pos=0.47, sloped, below] {$f^\star_1(x)$};
\addplot[color=cyan, thick, name path=func3, domain=0:20] {(x - 1)/(2 * x + 1)} node[pos=0.4, sloped, below] {$f^\star_2(x)$};
\addplot[color=orange, thick, name path=func3, domain=0:20] {(x - 1)/(0.5 * x + 1)} node[pos=0.4, sloped, above] {$f^\star_{1/2}(x)$};

% Constant function h_-(x) = -1
\addplot[color=black, thick, dashed, name path=func1, domain=0:20] {-1} node[pos=.2, sloped, below] {$h_{-}(x)$};

% Piecewise linear function h_+(x) = min(2x-1, 1)
\addplot[color=black, mark=*, mark size=1.5] coordinates {(0,-1)};

% Shaded area between functions
\addplot[color=gray, fill=gray!20, opacity=0.5, domain=0:10] fill between[of=func1 and func2];
\end{axis}
\end{tikzpicture}
}
% \vspace{-10.8cm}
 % Reduced from -2.5cm to -1.8cm to decrease space
    % \captionsetup{
    %     justification=centering,
    %     skip=-50pt,      % Space before caption
    %     position=bottom % Explicitly set position
    % }
    \caption{$f^\star_{n_r}$ for $n_r \in \{1/2, 1, 2\}$. }
    \label{fig:optimal_f}
    % \vspace{0.5cm}
\end{figure}

After demonstrating that a large panel of choices can lead to improved estimators, we focus on identifying the function $f$ that will result in the lowest variance, unbiased estimator. We give the resulting transform $f^*$ in the following proposition.

\begin{proposition}[Surrogate-Optimal improvement estimator] \label{optimal_est}
We set $n_r= n_A/n_B$. The function $f^*$ that minimizes the variance surrogate $S_f$ is defined as:
\begin{align}
    \forall x \in \mathbb{R}^+, \quad f^\star_{n_r}(x) = \frac{x - 1}{n_r x + 1}.
\end{align}
This function results in a simple, unbiased estimator ($f$ respects condition \eqref{eq:cond_1}) of the form:
\begin{align*}
    \hat{\mathcal{I}}_{f^\star_{n_r}} = \frac{1}{n_B} \sum_{j \in u_B} \sum_{t = 1}^{T_j} \frac{w_A(\tau^j_t) - 1}{n_r w_A(\tau^j_t) + 1} r^j_t + \frac{1}{n_A} \sum_{i \in u_A} \sum_{t = 1}^{T_i} \frac{1 - w_B(\tau^i_t)}{1 +  w_B(\tau^i_t)/n_r} r^i_t\,.
\end{align*}

\end{proposition}

\begin{proof}
The function $S_f$ we want to minimize decomposes over the expectation, and every trajectory $\tau_t$ can be linked to a one dimensional variable $f(w_A(\tau_t))$. This means that we can solve this minimization problem by looking for one variable $f(w_A(\tau_t))$ at a time, that sets the derivative to $0.$ Setting the derivative w.r.t $y = f(w_A(\tau_t))$ to $0$ gives:
\begin{align*}
    \frac{1}{n_B}f(w_A(\tau_t))  =  \frac{1}{n_A}\left(1 - w_B(\tau_t)\left[ 1 + f(w_A(\tau_t))\right] \right)\,,
\end{align*}
which is equivalent, after a few manipulations to:
\begin{align*}
    f(w_A(\tau_t))  =  \frac{w_A(\tau_t) - 1}{n_r w_A(\tau_t)  + 1}\,.
\end{align*}
\end{proof}
For all values of $n_r$, the function $f^\star_{n_r}$ is null when evaluated at $1$, i.e. $\forall n_r, f^\star_{n_r}(1) = 0$. This ensures that our estimator suffers no variance when the evaluated policies are identical, and will demonstrate substantial variance reduction when $\pi_A$ and $\pi_B$ behave similarly. Another good property that the optimal function encodes is its explicit dependence on the sample ratio between the two populations. This was not captured for instance by our previous construction even if it led to improved estimators. For instance, $n_A \gg n_B$, leads to a variance dominated by the term estimated with population $\mathcal{D}_B$. The function $f^\star_{n_r}$ reduces the importance of terms estimated by $\mathcal{D}_B$ and puts more mass on the terms estimated with $\mathcal{D}_A$. To better understand this behavior, we set $n_{U} = n_A + n_B$, $\hat{\beta} = n_A/n_{U}$, and define the mixture policy $\pi_{\hat{\beta}}$ for any trajectory $\tau_t$ by $\pi_{\hat{\beta}}(\tau_t) = \hat{\beta} \pi_A(\tau_t) + (1 - \hat{\beta}) \pi_B(\tau_t)$. We can rewrite our estimator:
\begin{align*}
    \hat{\mathcal{I}}_{f^\star_{n_r}} = \frac{1}{n_{U}} \sum_{m = 1}^{n_{U}} \sum_{t = 1}^{T_m} \frac{\pi_A(\tau^m_t) - \pi_B(\tau^m_t)}{\pi_{\hat{\beta}}(\tau^m_t)} r^m_t\,,
\end{align*}

reinterpreting our estimator as an Inverse Propensity Scoring (IPS) estimator of the improvement, leveraging the whole data $\mathcal{D}_U = \mathcal{D}_A \cup \mathcal{D}_B$, as it was collected by a mixture over trajectories. $\hat{\beta}$ controls the mixture weights and encodes implicitly the importance we give to both policies, approximately replicating the data generating process. Note that this estimator was already used in off-policy evaluation under multiple loggers within the contextual bandit framework \cite{BIPS}. Here, we prove its optimality under our construction for the general A/B testing problem, even with non-Markovian dependencies. Observe that the optimal function $f^*$ does not always reside in the region defined by condition \eqref{eq:cond_3} (it does when $n_r \ge 1$)  as seen in Figure~\ref{fig:optimal_f}. Indeed, \eqref{eq:cond_3} is only a sufficient condition to reduce the surrogate variance and the optimal function can reside outside the defined region to deal with the potential imbalance between the two populations.

\paragraph{\textbf{Asymptotic Normality and Decision Making.}}
For a fixed transform $f$, $\hat I_f$ is an average of independent user-level contributions from the two experimental populations. Therefore, under finite second moments, the central limit theorem gives asymptotic normality around $I(\pi_A,\pi_B)$. In practice, the variance is estimated as the empirical variance of these user-level contributions in each population, yielding asymptotically valid confidence intervals and one-sided lower confidence bounds necessary for decision making.

\subsection{In Practice: Propensities Misspecification}

\begin{figure}
    \centering

\resizebox{0.4\textwidth}{!}{%
    \begin{tikzpicture}
\begin{axis}[
    xlabel={$x$},
    ylabel={$f(x)$},
    xmin=-0.5, xmax=5,
    ymin=-1.3, ymax=1.2,
    samples=500,
    grid=major,
    legend pos=outer north east,
    ylabel style={yshift=-0.5cm},
    y label style={at={(axis description cs:-0.1,0.5)},rotate=270},
    yticklabels={,{$-1$},,,,{$1$}}
]

% \addplot[color=red, thick, name path=func3, domain=0:10] {-2 + 2 / (1 + exp(-x))} node[pos=0.2, sloped, above] {$g_1(x)$};
\addplot[color=orange, thick, name path=func4, domain=0:10] {-1/(2*x + 1)} node[pos=0.18, sloped, below] {$f^\star_{\Delta_1}(x)$};

\addplot[color=blue, thick, name path=func5, domain=0:10] {{(-1 + x*(1 - (x - 1)^2)) / (1 + x*(1 + (x - 1)^2))}} node[pos=0.4, sloped, above] {$f^\star_{\Delta_2}(x)$};

\addplot[color=teal, thick, name path=func6, domain=0:10] {{(-1 + x*(1 - (min( abs(ln(x+0.0001)), 1))^2)) / (1 + x*(1 + (min( abs(ln(x+0.0001)), 1))^2))}} node[pos=0.4, sloped, above] {$f^\star_{\Delta_3}(x)$};
% \addplot[color=orange, thick, name path=func3, domain=0:10] {min(pow(x,2)-1, 1)} node[pos=0.2, sloped, above] {};
\addplot[color=red, thick, name path=func3, domain=0:10] {(x-1)/(x + 1)} node[pos=0.3, sloped, below] {$f^\star_1(x)$};
% \addplot[color=red, thick, name path=func3, domain=0:10] {max(-1, min(0.5*(x-1), 1))} node[pos=0.2, sloped, above] {};

% Constant function h_-(x) = -1
\addplot[color=black, thick, dashed, name path=func1, domain=0:5] {-1} node[pos=.5, sloped, below] {$h_{-}(x)$};

% Piecewise linear function h_+(x) = min(2x-1, 1)
\addplot[color=black, thick, dashed, name path=func2, samples=500, domain=0:5] {min(2*x-1, 1)} node[pos=0.2, sloped, above] {$h_{+}(x)$};

\addplot[color=black, mark=*, mark size=1.7] coordinates {(0,-1)};
\addplot[color=black, mark=*, mark size=1.7] coordinates {(1,0)};

% Shaded area between functions
\addplot[color=gray, fill=gray!20, opacity=0.5, domain=0:10] fill between[of=func1 and func2];
\end{axis}
\end{tikzpicture}
}
    \caption{Optimal $f$ for $\lambda =1$ and  for different noise models $\Delta$. $\Delta_1: x \rightarrow 1$, $\Delta_2: x\rightarrow |x - 1|$, $\Delta_3: x\rightarrow \min(|\log x|, 1)$.}
    \label{fig:robust}
\end{figure}

In many practical applications, the propensities of the tested policies are not directly accessible. In such cases, we estimate them from data, obtaining $\hat{\pi}_A(a\mid s)$ and $\hat{\pi}_B(a\mid s)$ for all actions $a$ and states $s$, and then apply our method to construct a better-behaved estimator. However, even if $\hat{\pi}_A$ and $\hat{\pi}_B$ are unbiased estimators of the true propensities, the resulting estimator can still be biased, since importance weights involve \emph{ratios} of propensities. This suggests that the variance-optimal choice of $f$ derived under known propensities may no longer be optimal when propensities are estimated. We therefore study the sensitivity of the general $f$-A/B testing estimator to misspecification of the importance weights. Consider a single interaction $(s,a)$ (or a single-step trajectory $\tau=\{(s,a)\}$). We model propensity ratio misspecification through the following multiplicative perturbation:
\begin{align*}
\frac{\hat{\pi}_B(a\mid s)}{\hat{\pi}_A(a\mid s)}
=
\frac{\pi_B(a\mid s)}{\pi_A(a\mid s)}
\bigl(1+\delta\,\Delta(a,s)\bigr),
\end{align*}
with $\Delta(a,s) \ge 0$ any function (can depend on $a$ and $s$) and $\delta \in \mathbb{R}$ a small bias perturbation. $\delta = 0$ ensures a null bias. For $\delta \rightarrow 0$, the misspefication for any trajectory $\tau_t = \{a_1, s_1, \cdots , a_t, s_t \}$ gives:
\begin{align*}
     \hat{w}_A(\tau_t) &= w_A(\tau_t) \prod_{\ell = 1}^t(1 + \delta\Delta(a_\ell,s_\ell)) \approx w_A(\tau_t)\left(1 + \delta \Delta(\tau_t)\right)\,,
\end{align*}
with $\Delta(\tau_t) = \sum_{\ell =1}^t\Delta(a_\ell,s_\ell)$. With another Taylor expansion argument, $\hat{w}_B(\tau_t)$ is recovered by:
\begin{align*}
    \hat{w}_B(\tau_t) = \frac{1}{\hat{w}_A(\tau_t)} &\approx w_B(\tau_t)\left(1  - \delta \Delta(\tau_t)) \right) \,.
\end{align*}
Different choices of $\Delta(a,s)$ will recover different noise models, for example $\Delta(a,s) = 1$, recovers a classical multiplicative noise and $\Delta(a,s) = |\pi_B(a|s) /\pi_A(a|s) - 1|$ recovers a noise that scales with how far the two tested policies are. We start by writing down the $f$-regularised estimator with misspecified weights:
\begin{align*}
    \tilde{\mathcal{I}}_{f} = &\frac{1}{n_B} \sum_{j \in u_B}\sum_{t = 1}^{T_j} f(\hat{w}_A(\tau^j_t)) r^j_t \\ &+ \frac{1}{n_A} \sum_{i \in u_A} \sum_{t = 1}^{T_i} \left(1 - \hat{w}_B(\tau^i_t)\left[ 1 + f(\hat{w}_A(\tau^i_t))\right] \right) r^i_t\,,
\end{align*}
with the condition $f(0) = -1$. As we are using misspecified weights, the unbiased guarantees on our estimator do not hold anymore. As $\delta \rightarrow 0$, we can invoke a taylor argument and quantify the bias to the first order with:
\begin{align*}
    b_\delta(\Delta, f) = \delta \,\mathbb{E}_{\nu(\pi_B)}\left[\sum_{t=1}^T \Delta(\tau_t)(1 + f(w_A(\tau_t)))r_t\right] \,.
\end{align*}
This bias is naturally null when $\delta$ or $\Delta$ are null. Controlling this bias means that we want $f$ to be close to $-1$ when trajectory noise level $\Delta(\tau_t)$ is substantial. Finding $f$ that minimizes the MSE is equivalent to finding an $f$ that optimally minimizes the bias-variance tradeoff. In the small $\delta$ regime, using Jensen's inequality, Cauchy-Schwarz and the variance surrogate, we get: 
\begin{align*}
    \text{MSE}(\tilde{\mathcal{I}}_{f}) \le  \delta^2 T B_{\Delta}(f)  + S_f \,,
\end{align*}
with $B_{\Delta}(f) = \mathbb{E}_{\nu(\pi_B)}\left[\sum_{t=1}^T \left(\Delta(\tau_t)(1 + f(w_A(\tau_t))) r_t\right)^2\right]$. This results in the following selection for $f$, To define robust estimators, we look for the function $f$ that minimizes the following tradeoff:
\begin{align*}
    S_{\lambda, \Delta}(f) = \lambda T B_{\Delta}(f) + S_f\,,
\end{align*}
with $\lambda \ge 0$. Appendix~\ref{app:misspecification} details the construction of the objective. Solving for the function $f$ that minimizes this tradeoff, we get the following proposition.
\begin{proposition}[Misspecficiation-Aware Optimal Estimator] Let $\Delta$ be the weights misspecification model used and $\lambda > 0$ the bias-variance tradeoff parameter. The function $f^\star_{\lambda, \Delta}$ that minimizes the bias-variance tradeoff writes for any trajectory $\tau_t$:
\begin{align*}
    f^\star_{\lambda, \Delta}(w(\tau_t)) = \frac{(1 - \lambda T n_A \Delta(\tau_t)^2)w(\tau_t) - 1}{(n_r + \lambda T n_A \Delta(\tau_t)^2)w(\tau_t) + 1}\,.
\end{align*}
\end{proposition}
\begin{proof}
Fix a prefix $\tau_t$ and write $w=w_A(\tau_t)$, $w_B(\tau_t)=1/w$, and $\Delta=\Delta(\tau_t)$. Consider the pointwise objective (dropping terms independent of $f(w)$)
\[
S_{\lambda,\Delta}(f)
=
\lambda\,T\,\Delta^2\bigl(1+f(w)\bigr)^2
+\frac{1}{n_B}f(w)^2
+\frac{1}{n_A}\Bigl(1-w_B(\tau_t)\bigl(1+f(w)\bigr)\Bigr)^2.
\]
Let $y=f(w)$. Differentiating w.r.t.\ $y$ and setting to zero gives
\[
\lambda T\Delta^2(1+y)+\frac{1}{n_B}y
=
\frac{1}{n_A}\Bigl(1-w_B(\tau_t)(1+y)\Bigr),
\]
which expands to
\[
\Bigl(\lambda T\Delta^2+\frac{w_B}{n_A}\Bigr)
+
y\Bigl(\lambda T\Delta^2+\frac{1}{n_B}+\frac{w_B}{n_A}\Bigr)
=
\frac{1}{n_A}.
\]
Solving for $y$ yields
\[
y
=
\frac{\frac{1}{n_A}-\lambda T\Delta^2-\frac{w_B}{n_A}}
{\lambda T\Delta^2+\frac{1}{n_B}+\frac{w_B}{n_A}}
=
\frac{\frac{1}{n_A}-\lambda T\Delta^2-\frac{1}{n_A w}}
{\lambda T\Delta^2+\frac{1}{n_B}+\frac{1}{n_A w}}.
\]
Multiplying numerator and denominator by $n_A w$ gives
\[
y
=
\frac{w-1-\lambda T\Delta^2\,n_A w}{\lambda T\Delta^2\,n_A w+\frac{n_A}{n_B}w+1}.
\]
Finally, $n_r=n_A/n_B$ yields:
\[
f^\star_{\lambda,\Delta}(w(\tau_t))
=
\frac{\bigl(1-\lambda T n_A\Delta(\tau_t)^2\bigr)w(\tau_t)-1}
{\bigl(n_r+\lambda T n_A\Delta(\tau_t)^2\bigr)w(\tau_t)+1}.
\]
Since the objective is a strictly convex quadratic in $y$, this stationary point is the unique minimizer.
\end{proof}
This function recovers $f^\star_{n_r}$ when $\lambda = 0$ and recovers the difference in means estimator ($f = -1$) when $\lambda \rightarrow \infty$. For $\lambda > 0$, it inherits some of properties of $f^\star_{n_r}$ as it depends on the data imbalance $n_r$, its value at $0$ is still $f(0) = -1$, which means that using this in the true importance weights case should also provide an unbiased estimator. When $n_r = 1$, Figure~\ref{fig:robust} shows that for different choices of $\Delta$, the resulting function remains within the surrogate improvement region. As a consequence, any estimator is expected to improve upon the variance of the standard difference-in-means estimator. Finally, depending on the noise model, the estimator may lose the property of returning zero when the two policies are identical. This property is preserved only when the induced noise along the trajectory is itself null when the propensities match. 

\paragraph{\textbf{On the noise model.}}
The choice of $\Delta$ should reflect how propensity misspecification is expected to appear in the application. When propensities are learned by maximum likelihood, a logarithmic noise model is particularly natural. This is discussed in Appendix~\ref{app:noise_model}. This motivates the clipped logarithmic choice $\Delta(x)=\min(|\log x|,1)$, which is symmetric, vanishes when the policies agree, grows smoothly as they separate, and avoids excessive sensitivity to extreme ratios.

\subsection{Cases of marginal improvement}\label{sec:limitations_main}

\paragraph{\textbf{Perfectly distinct policies.}} Recall that the motivation of our family of estimators is to exploit the similarities between the two tested policies, achieving a variance reduction on states for which the two policies have the same behavior. In the special case where the two policies never play the same actions, our newly introduced family of estimators will default to the difference estimator. Precisely, if policies $\pi_A$ and $\pi_B$ have a completely distinct support, meaning that for all $s \in \mathcal{S}$ and $a \in \mathcal{A}$, $\pi_A(a|s)\pi_B(a|s) = 0$, then the importance weight $w_B(\tau)$ of $\pi_B$ computed for any trajectory $\tau$ played with $\pi_A$ will be $w_B(\tau) = 0$ and similarly, the importance weight $w_A(\tau)$ of $\pi_A$ computed for any trajectory $\tau$ played with $\pi_B$ will be $w_A(\tau) = 0$. This means that for all terms will be evaluated at $0$, and as $f(0) = - 1$, our estimator will default to the difference estimator giving $\hat{\mathcal{I}}_{f} = \hat{\mathcal{I}}\,$, no matter the function $f$ used.
% \Dave{
% \begin{align*}
%     \hat{\mathcal{I}}_{f}(\pi_A, \pi_B) &= \frac{1}{|n_B|} \left(\sum_{j \in n_B} \sum_{t = 1}^{T_j} f(0) r^j_t \right) + \frac{1}{|n_A|} \left( \sum_{i \in n_A} \sum_{t = 1}^{T_i}  r^i_t \right) \\
%     &= \hat{\mathcal{I}}(\pi_A, \pi_B)\,.  
% \end{align*}
% }

\paragraph{\textbf{Long horizon rewards.}} In some applications, the reward signal can only be measured after multiple interactions. One can think of a sales signal that happens after multiple recommendations. Formally, let us suppose that the observed reward can only be positive after $T_0 \gg 1$ interactions. In this scenario, even if the two tested policies are only slightly different, their propensities of playing the same trajectory will drift apart, becoming less and less similar as the trajectory grows. The following example details this phenomenon. Let $0 < \epsilon \le 1$, suppose that our action space is binary $\mathcal{A} = \{a^{-}, a^{+}\}$, and take the simple example of testing the two following policies: for all states $s$, $\pi_A(a^+|s) = 0$ and $\pi_B(a^+|s) = \epsilon.$ These policies can be made very close by varying $\epsilon \rightarrow 0$. Let us suppose that the trajectory stops when we get a positive reward, and let that time always be $T_0$. The first term of our estimator is an average over $\pi_B$. The probability of the event $\{w_A(\tau_{T_0}) > 0 \}$ is equal to $(1 - \epsilon)^{T_0}$, the probability of $\pi_B$ playing $T_0$ consecutive $a^+$. This probability decays exponentially to $0$, meaning that for $T_0 \gg 1$, the first term of our estimator will consist of importance weights of $w_A(\tau_{T_0}) = 0$. Similarly, the second term of the estimator is an average over trajectories generated by $\pi_A$, which is deterministic in our case and results in importance weights of $w_B(\tau_{T_0}) = (1 - \epsilon)^{T_0}$. For large $T_0 \gg 1$, this converges to $0$, leading to an estimator evaluated on null importance weights. As $f(0) = -1$, our family of estimators defaults back to the difference-in-means estimator when dealing with long-horizon rewards, and leads to no variance reduction. However, this limitation can be addressed with advanced importance weighting techniques that exploit the sequential structure of the problem \cite{cond_is, curse_long, long_hor_cond}. Such improvements, while promising, lie beyond the scope of this work.

\begin{figure*}
    \centering
    \includegraphics[width=\linewidth]{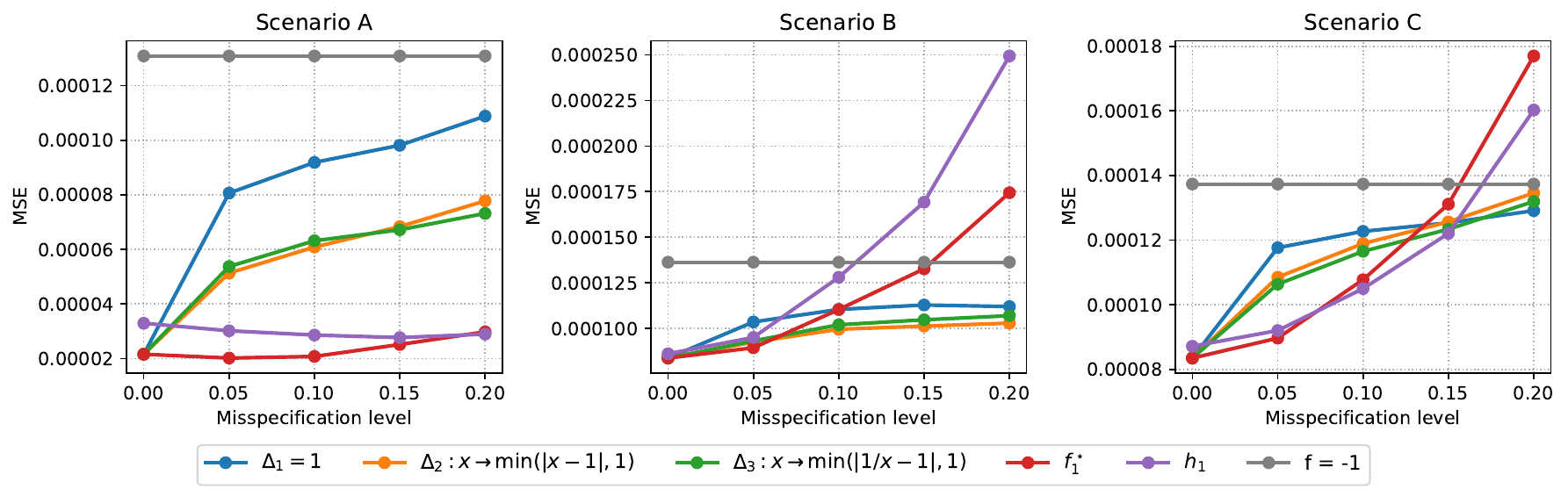}
    \caption{MSE of different estimators across different scenarios and misspecification levels. Misspecification-aware estimators are robust and enjoy a better MSE than Difference-in-means even for moderate misspecification levels.}
    \Description{Three-panel plot showing mean squared error as a function of propensity misspecification level for scenarios A, B, and C. The difference-in-means estimator remains stable, while non-robust importance-weighted estimators degrade as misspecification increases. The misspecification-aware estimators maintain lower MSE across the scenarios.}
    \label{fig:mse}
\end{figure*}

\paragraph{\textbf{Strong propensities misspecification.}} Our family of estimators extends to practical settings in which propensities must be estimated, as discussed in the previous section. In principle, if the induced bias due to propensity misspecification were large, controlling it would require choosing a large regularization parameter~$\lambda$, potentially in the limit $\lambda \to \infty$. In this regime, the proposed misspecification-aware estimators reduce to the difference-in-means estimator, since $f_{\lambda,\Delta} \to -1$ as $\lambda \to \infty$. In practice, this extreme scenario is unlikely to arise: commonly used propensity score estimators exhibit controllable and diminishing bias as the sample size~$n$ increases \citep{kallusdouble2020}.

\section{Experiments}

\begin{figure}[ht]
    \centering
    \includegraphics[width=\linewidth]{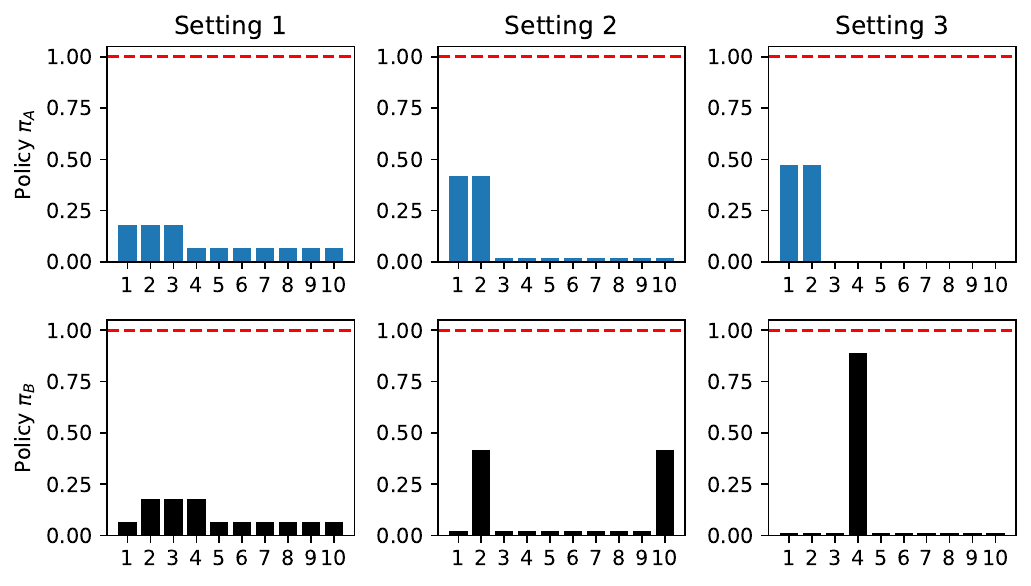}
    \caption{Policies used in experiment (1).}
    \label{fig:pol_10}
\end{figure}

\paragraph{\textbf{(1) Overlapping policies and data imbalance.}} We want to empirically validate the behavior of our family of estimators and quantify the variance reduction $v_r(f) = \text{var}(\hat{\mathcal{I}})/\text{var}(\hat{\mathcal{I}_f})$ achieved when varying the overlap between the two A/B tested policies $\pi_A$ and $\pi_B$, and varying $n_r = n_A/n_B$, the ratio of users allocated to each policy. For these experiments, we focus on a simplified setting and simulate a bandit problem ($T = 1$ and $\lvert S\rvert = 1$) with a discrete action space $\mathcal{A}$ of size $\lvert\mathcal{A}\rvert = 10$ and binary, noisy sparse rewards reminiscent of online decision systems scenarios. We measure the distance $d(\pi_A, \pi_B)$ between the two policies as the average of variances of importance weights $w_A$ and $w_B$, varying from $0$ (identical policies) to very large (distinct policies). We use 3 settings with different policy overlaps (Figure~\ref{fig:pol_10}). Results of these experiments are summarized in Table~\ref{tab:v_gain_bandit} and match our theoretical findings. The variance reduction is substantial when the policies are close, and it is marginal when the tested policies are really far. This reduction is also accentuated by the presence of imbalance in user allocation ($n_r \neq 1$). Intuitively, the difference estimator suffers the variance of the smallest population, while our off-policy estimators, especially the optimal $f^\star_{n_r}$, efficiently combine the two populations, achieving accurate estimation even when $n_r \neq 1$. Appendix~\ref{app:detailed_experiments} details the experimental setup and provides more comprehensive results.

\begin{table}
    \centering
    \caption{Variance reduction $v_r(f)$ w.r.t. $d(\pi_A,\pi_B)$.}
    \label{tab:v_gain_bandit}
    \begin{tabular}{ccc||ccc}
        \toprule
        Setting & $d(\pi_A,\pi_B)$ & $n_r$ & $v(h_1)$ & $v(f^\star_1)$ & $v(f^\star_{n_r})$ \\
        \midrule

        \multirow{3}{*}{\textbf{(1)}} & \multirow{3}{*}{$\sim 1$} 
        & $1/4$ & \underline{$26.53$} & $19.01$ & $\boldsymbol{27.52}$ \\
        & & $1$ & $13.29$ & $\boldsymbol{19.05}$ & $\boldsymbol{19.05}$ \\
        & & $4$ & $8.87$ & \underline{$19.10$} & $\boldsymbol{27.44}$ \\

        \midrule

        \multirow{3}{*}{\textbf{(2)}} & \multirow{3}{*}{$\sim 10$}
        & $1/4$ & \underline{$2.78$} & $2.72$ & $\boldsymbol{3.05}$ \\
        & & $1$ & $2.69$ & $\boldsymbol{2.76}$ & $\boldsymbol{2.76}$ \\
        & & $4$ & $2.60$ & \underline{$2.69$} & $\boldsymbol{3.02}$ \\

        \midrule

        \multirow{3}{*}{\textbf{(3)}} & \multirow{3}{*}{$\sim 10^2$}
        & $1/4$ & \underline{$1.13$} & $1.12$ & $\boldsymbol{1.20}$ \\
        & & $1$ & $1.11$ & $\boldsymbol{1.13}$ & $\boldsymbol{1.13}$ \\
        & & $4$ & $1.08$ & $\boldsymbol{1.15}$ & $\boldsymbol{1.15}$ \\

        \bottomrule
    \end{tabular}
\end{table}

\begin{figure*}
    \centering
    \includegraphics[width=\linewidth]{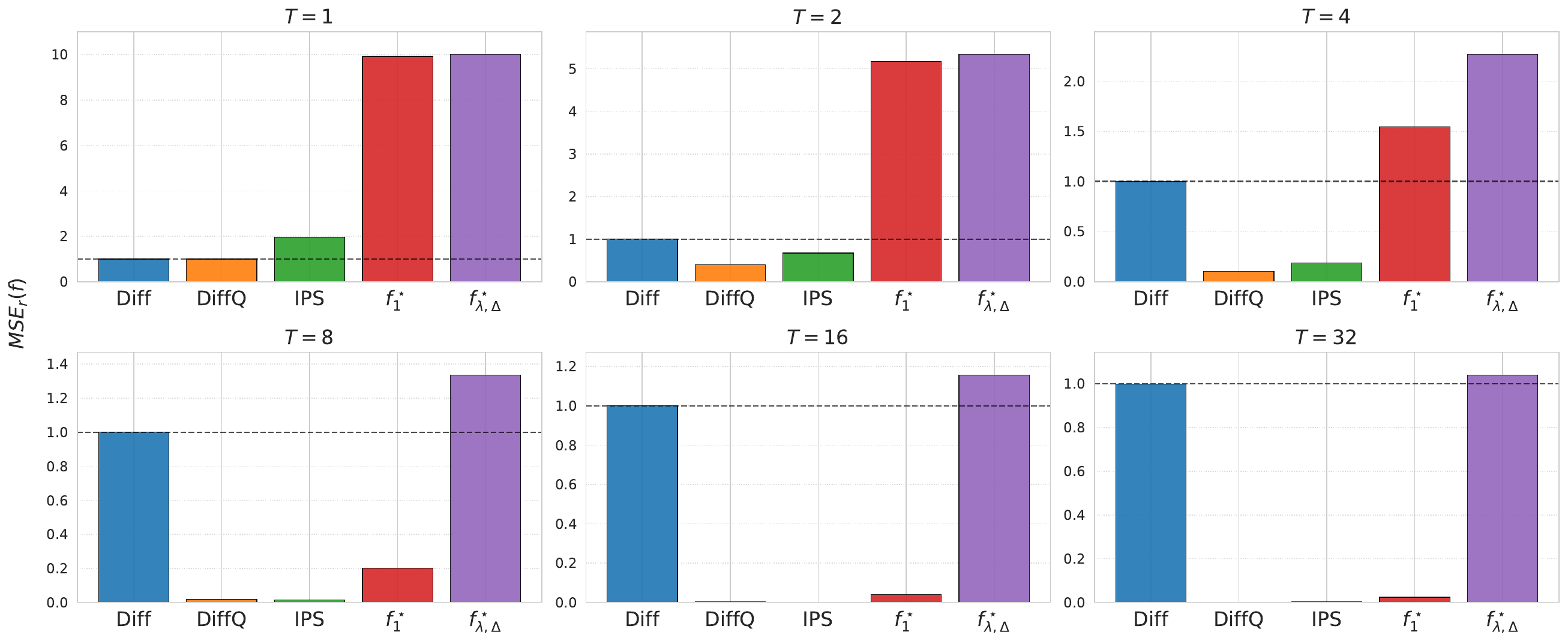}
    \caption{Relative MSE Reduction across $T \in \{1, 2, 4, 8, 16, 32\}$. \texttt{IPS} and \texttt{DiffQ} fail for moderate $T$. $f^\star_1$ exhibits lower MSE for small $T$ and fails at larger $T$. The robust estimator with $f^\star_{\lambda, \Delta}$ exhibits better MSE for all $T$.}
    \label{fig:abtest}
\end{figure*}

\paragraph{\textbf{(2) Sensitivity to Weight Misspecification.}}
We empirically evaluate the robustness of the proposed misspecification-aware estimators to errors in propensity estimation and compare their mean squared error (MSE) to that of alternative estimators. We consider a simplified bandit problem with a discrete action space $\mathcal{A}$ of size $\lvert\mathcal{A}\rvert = 10$ and set $n_A = n_B$. We study three scenarios: \emph{(A)} overlapping policies, \emph{(B)} $\pi_A$ more concentrated than $\pi_B$, and \emph{(C)} $\pi_B$ more concentrated than $\pi_A$. Propensity misspecification is induced by perturbing the true propensities: for any $a \in \mathcal{A}$, $\hat{\pi}_A(a) = (1 - \sigma)\,\pi_A(a) + \sigma/\lvert\mathcal{A}\rvert,
$ and similarly for $\hat{\pi}_B$, where $\sigma \ge 0$ controls the misspecification level. We evaluate our misspecification-aware optimal functions under three noise models, with the regularization parameter $\lambda$ increasing with~$\sigma$. Figure~\ref{fig:mse} reports the results. The difference-in-means estimator ($f=-1$) exhibits stable MSE, as it does not rely on importance weights. In contrast, in the more challenging scenarios (B and~C), the MSE of $h_1$ and $f^\star$ grows rapidly with $\sigma$, in some cases exceeding that of the difference-in-means estimator, indicating a lack of robustness to misspecification. By contrast, the proposed robust estimators consistently outperform the difference-in-means baseline across all scenarios, even under substantial misspecification. Note that $\Delta_1 = 1$ is more conservative than the other choices, since it does not vanish even when the two policies coincide.

\paragraph{\textbf{(3) A Realistic Recommendation Problem.}}
We simulate multiple A/B testing scenarios in a recommendation environment. We construct a Markov Chain to model user boredom, wherein repeatedly recommending the same item reduces user engagement over time. Such environments incentivize diffuse policies that diversify recommendations to maintain user interest. We set $|\mathcal{A}| = d = 10$. For any $t$, the user state is denoted by $\boldsymbol{s}_t \in \mathbb{R}^d$, and the non-Markovian Decision Process evolves as : 
$$\boldsymbol{s}_{t+1} = \rho \boldsymbol{s}_0 + (1-\rho)\left(\boldsymbol{s}_t + \sigma \boldsymbol{\epsilon} - \beta_a \odot \boldsymbol{s}_t\right) \,,$$ where $r_t \sim B(y_t), \quad y_t \propto \langle \beta_a, \boldsymbol{s}_t\rangle\,$ 
$\rho$ controls the markovian assumption, $\boldsymbol{\epsilon} \in \mathbb{R}^d$ is Gaussian noise, $\sigma$ controls its scale, $\beta_a \in \mathbb{R}^d$ is the parameter vector associated with action $a$, $\odot$ the element-wise multiplication and $B(y_t)$ denotes a Bernoulli distribution with mean $y_t$. We restrict $\boldsymbol{s}_{t} \in [0, 1]^d$ and $y_t \in [0, 1]$ by clipping and use $\rho = 0.25$ to break the markovian assumption. We collect $N=1000$ interactions under each policy $\pi_A$ and $\pi_B$, yielding balanced datasets $\mathcal{D}_A$ and $\mathcal{D}_B$. Propensities of both policies are \emph{learned} from the logged data, placing us in a realistic A/B test setting. We evaluate several A/B testing estimators: \texttt{Diff} (difference-in-means), \texttt{DiffQ} (difference-in-$Q$ values) from \cite{farias_dq} designed for Markovian interference, \texttt{IPS} (inverse propensity scoring; Eq.~\eqref{eq:naive_ips}), and our regularized estimators using both $f^\star_{\lambda,\Delta}$  ($\Delta(x)=\min\{|\log x|,1\}$, $\lambda=0.5$) and $f^\star_1$. Experiments are conducted over horizons $T \in \{1,2,4,8,16,32\}$, with each configuration repeated $500$ times to estimate mean squared error. Results are summarized in Figure~\ref{fig:abtest}, which reports the relative MSE reduction $\mathrm{MSE}_r(f) = \mathrm{MSE}(\hat{\mathcal{I}}) / \mathrm{MSE}(\hat{\mathcal{I}}_f)$,
normalized by the difference-in-means estimator. While \texttt{DiffQ} is unbiased, it suffers from high variance and is consistently outperformed by \texttt{Diff}. Naive \texttt{IPS} benefits from policy similarity at $T=1$ but becomes unstable as the horizon grows. The variance-optimal estimator $f^\star_1$ substantially improves upon \texttt{Diff} for moderate horizons, but its performance degrades sharply for large~$T$ due to compounding propensity estimation error. In contrast, our misspecification-aware estimators consistently outperform \texttt{Diff}, providing significant variance reduction at short horizons and stable, more modest, gains at longer ones.

\section{Conclusion}
Classical A/B testing remains the gold standard for comparing decision systems. In this work, we showed that its core estimation strategy can be improved by exploiting similarities between the tested policies. We introduced a family of importance-weighted A/B testing estimators that remain compatible with the standard experimental protocol, reduce to the difference-in-means estimator when no useful overlap is present, and achieve substantial mean squared error reductions when the policies are similar.

Our results suggest that off-policy ideas can be used not only as an alternative to online experimentation, but also as a tool to make A/B testing itself more statistically efficient. Several extensions remain promising directions for future work, including doubly robust estimators \citep{dudik14doubly} to further reduce reward variance, and methods that exploit the sequential structure of user trajectories to address the challenges posed by long horizons \citep{curse_long, cond_is, long_hor_cond}.

\newpage
\bibliographystyle{ACM-Reference-Format}
\balance
\bibliography{bib}

@article{kallusdouble2020,
author = {Kallus, Nathan and Uehara, Masatoshi},
title = {Double reinforcement learning for efficient off-policy evaluation in Markov decision processes},
year = {2020},
issue_date = {January 2020},
publisher = {JMLR.org},
volume = {21},
number = {1},
issn = {1532-4435},
journal = {J. Mach. Learn. Res.},
month = jan,
articleno = {167},
numpages = {63}
}

@article{shi2021time,
  title={Dynamic Causal Effects Evaluation in A/B Testing with a Reinforcement Learning Framework},
  author={Shi, Chengchun and Wang, Xiaoyu and Luo, Shikai and Zhu, Hongtu and Ye, Jieping and Song, Rui},
  journal={Journal of the American Statistical Association},
  volume={accepted},
  year={2021}
}

@misc{chen2024experimenting,
      title={Experimenting on Markov Decision Processes with Local Treatments}, 
      author={Shuze Chen and David Simchi-Levi and Chonghuan Wang},
      year={2024},
      eprint={2407.19618},
      archivePrefix={arXiv},
      primaryClass={stat.ME},
      url={https://arxiv.org/abs/2407.19618}, 
}

@inproceedings{farias_dq,
author = {Farias, Vivek and Li, Hao and Peng, Tianyi and Ren, Xinyuyang and Zhang, Huawei and Zheng, Andrew},
title = {Correcting for Interference in Experiments: A Case Study at Douyin},
year = {2023},
isbn = {9798400702419},
publisher = {Association for Computing Machinery},
address = {New York, NY, USA},
url = {https://doi.org/10.1145/3604915.3608808},
doi = {10.1145/3604915.3608808},
booktitle = {Proceedings of the 17th ACM Conference on Recommender Systems},
pages = {455–466},
numpages = {12},
location = {Singapore, Singapore},
series = {RecSys '23}
}

@inproceedings{farias_neurips,
author = {Farias, Vivek and Li, Andrew A. and Peng, Tianyi and Zheng, Andrew},
title = {Markovian interference in experiments},
year = {2022},
isbn = {9781713871088},
publisher = {Curran Associates Inc.},
address = {Red Hook, NY, USA},
booktitle = {Proceedings of the 36th International Conference on Neural Information Processing Systems},
articleno = {39},
numpages = {15},
location = {New Orleans, LA, USA},
series = {NIPS '22}
}

@inproceedings{BIPS,
author = {Agarwal, Aman and Basu, Soumya and Schnabel, Tobias and Joachims, Thorsten},
title = {Effective Evaluation Using Logged Bandit Feedback from Multiple Loggers},
year = {2017},
isbn = {9781450348874},
publisher = {Association for Computing Machinery},
address = {New York, NY, USA},
url = {https://doi.org/10.1145/3097983.3098155},
doi = {10.1145/3097983.3098155},
booktitle = {Proceedings of the 23rd ACM SIGKDD International Conference on Knowledge Discovery and Data Mining},
pages = {687–696},
numpages = {10},
location = {Halifax, NS, Canada},
series = {KDD '17}
}

@article{agarwal2016multiworld,
  title={Multiworld Testing Decision Service: A System for Experimentation, Learning, And Decision-Making},
  author={Agarwal, Alekh and Bird, Sarah and Cozowicz, Markus and Dud{\i}k, Miro and Hoang, Luong and Langford, John and Li, Lihong and Melamed, Dan and Oshri, Gal and Sen, Siddhartha and others},
  journal={Whitepaper of Microsoft},
  pages={1--40},
  year={2016}
}

@inproceedings{beygelzimer2011contextual,
  title={Contextual bandit algorithms with supervised learning guarantees},
  author={Beygelzimer, Alina and Langford, John and Li, Lihong and Reyzin, Lev and Schapire, Robert},
  booktitle={Proceedings of the Fourteenth International Conference on Artificial Intelligence and Statistics},
  pages={19--26},
  year={2011},
  organization={JMLR Workshop and Conference Proceedings}
}

@article{bottou2013counterfactual,
  title={Counterfactual reasoning and learning systems: The example of computational advertising.},
  author={Bottou, L{\'e}on and Peters, Jonas and Qui{\~n}onero-Candela, Joaquin and Charles, Denis X and Chickering, D Max and Portugaly, Elon and Ray, Dipankar and Simard, Patrice and Snelson, Ed},
  journal={Journal of Machine Learning Research},
  volume={14},
  number={11},
  year={2013}
}

@inproceedings{joachims2018deep,
  title={Deep learning with logged bandit feedback},
  author={Joachims, Thorsten and Swaminathan, Adith and De Rijke, Maarten},
  booktitle={International Conference on Learning Representations},
  year={2018}
}

@inproceedings{swaminathan2015counterfactual,
  title={Counterfactual risk minimization: Learning from logged bandit feedback},
  author={Swaminathan, Adith and Joachims, Thorsten},
  booktitle={International Conference on Machine Learning},
  pages={814--823},
  year={2015},
  organization={PMLR}
}

@inproceedings{kohavi2012trustworthy,
  title={Trustworthy online controlled experiments: Five puzzling outcomes explained},
  author={Kohavi, Ron and Deng, Alex and Frasca, Brian and Longbotham, Roger and Walker, Toby and Xu, Ya},
  booktitle={Proceedings of the 18th ACM SIGKDD international conference on Knowledge discovery and data mining},
  pages={786--794},
  year={2012}
}

@inproceedings{kohavi2013online,
  title={Online controlled experiments at large scale},
  author={Kohavi, Ron and Deng, Alex and Frasca, Brian and Walker, Toby and Xu, Ya and Pohlmann, Nils},
  booktitle={Proceedings of the 19th ACM SIGKDD international conference on Knowledge discovery and data mining},
  pages={1168--1176},
  year={2013}
}

@InProceedings{pmlr-v162-wan22b,
  title = 	 {Safe Exploration for Efficient Policy Evaluation and Comparison},
  author =       {Wan, Runzhe and Kveton, Branislav and Song, Rui},
  booktitle = 	 {Proceedings of the 39th International Conference on Machine Learning},
  pages = 	 {22491--22511},
  year = 	 {2022},
  editor = 	 {Chaudhuri, Kamalika and Jegelka, Stefanie and Song, Le and Szepesvari, Csaba and Niu, Gang and Sabato, Sivan},
  volume = 	 {162},
  series = 	 {Proceedings of Machine Learning Research},
  month = 	 {17--23 Jul},
  publisher =    {PMLR}
}

@inproceedings{m2,
author = {Wang, Yu and Gupta, Somit and Lu, Jiannan and Mahmoudzadeh, Ali and Liu, Sophia},
title = {On Heavy-user Bias in A/B Testing},
year = {2019},
isbn = {9781450369763},
publisher = {Association for Computing Machinery},
address = {New York, NY, USA},
booktitle = {Proceedings of the 28th ACM International Conference on Information and Knowledge Management},
pages = {2425–2428},
numpages = {4},
location = {Beijing, China},
series = {CIKM '19}
}

@inproceedings{diversity,
author = {Wilhelm, Mark and Ramanathan, Ajith and Bonomo, Alexander and Jain, Sagar and Chi, Ed H. and Gillenwater, Jennifer},
title = {Practical Diversified Recommendations on YouTube with Determinantal Point Processes},
year = {2018},
isbn = {9781450360142},
publisher = {Association for Computing Machinery},
address = {New York, NY, USA},
url = {https://doi.org/10.1145/3269206.3272018},
doi = {10.1145/3269206.3272018},
booktitle = {Proceedings of the 27th ACM International Conference on Information and Knowledge Management},
pages = {2165–2173},
numpages = {9},
location = {Torino, Italy},
series = {CIKM '18}
}

@inproceedings{m1,
author = {Johari, Ramesh and Koomen, Pete and Pekelis, Leonid and Walsh, David},
title = {Peeking at A/B Tests: Why it matters, and what to do about it},
year = {2017},
isbn = {9781450348874},
publisher = {Association for Computing Machinery},
address = {New York, NY, USA},
url = {https://doi.org/10.1145/3097983.3097992},
doi = {10.1145/3097983.3097992},
booktitle = {Proceedings of the 23rd ACM SIGKDD International Conference on Knowledge Discovery and Data Mining},
pages = {1517–1525},
numpages = {9},
location = {Halifax, NS, Canada},
series = {KDD '17}
}

@inproceedings{search,
author = {Li, Lihong and Chu, Wei and Langford, John and Wang, Xuanhui},
title = {Unbiased offline evaluation of contextual-bandit-based news article recommendation algorithms},
year = {2011},
isbn = {9781450304931},
publisher = {Association for Computing Machinery},
address = {New York, NY, USA},
url = {https://doi.org/10.1145/1935826.1935878},
doi = {10.1145/1935826.1935878},
booktitle = {Proceedings of the Fourth ACM International Conference on Web Search and Data Mining},
pages = {297–306},
numpages = {10},
location = {Hong Kong, China},
series = {WSDM '11}
}

@inproceedings{pol_conv,
author = {Sachdeva, Noveen and Wang, Lequn and Liang, Dawen and Kallus, Nathan and McAuley, Julian},
title = {Off-Policy Evaluation for Large Action Spaces via Policy Convolution},
year = {2024},
isbn = {9798400701719},
publisher = {Association for Computing Machinery},
address = {New York, NY, USA},
url = {https://doi.org/10.1145/3589334.3645501},
doi = {10.1145/3589334.3645501},
booktitle = {Proceedings of the ACM Web Conference 2024},
pages = {3576–3585},
numpages = {10},
location = {Singapore, Singapore},
series = {WWW '24}
}

@book{ sutton98reinforcement,
  author = "Richard Sutton and Andrew Barto",
  title = "Reinforcement Learning: An Introduction",
  publisher = "MIT Press",
  address = "Cambridge, MA",
  year = "1998"
}

@inproceedings{
sakhi2024logarithmicsmoothingpessimisticoffpolicy,
title={Logarithmic Smoothing for Pessimistic Off-Policy Evaluation, Selection and Learning},
author={Otmane Sakhi and Imad Aouali and Pierre Alquier and Nicolas Chopin},
booktitle={The Thirty-eighth Annual Conference on Neural Information Processing Systems},
year={2024},
url={https://openreview.net/forum?id=zLClygeRK8}
}

@article{ dudik14doubly,
  author = "Miroslav Dudik and Dumitru Erhan and John Langford and Lihong Li",
  title = "Doubly Robust Policy Evaluation and Optimization",
  journal = "Statistical Science",
  volume = "29",
  number = "4",
  pages = "485-511",
  year = "2014"
}

@misc{in_out,
      title={Variance reduction combining pre-experiment and in-experiment data}, 
      author={Zhexiao Lin and Pablo Crespo},
      year={2024},
      eprint={2410.09027},
      archivePrefix={arXiv},
      primaryClass={stat.ME},
      url={https://arxiv.org/abs/2410.09027}, 
}

@InProceedings{alpha_test,
  title = 	 {On Validation and Planning of An Optimal Decision Rule with Application in Healthcare Studies},
  author =       {Cai, Hengrui and Lu, Wenbin and Song, Rui},
  booktitle = 	 {Proceedings of the 37th International Conference on Machine Learning},
  pages = 	 {1262--1270},
  year = 	 {2020},
  editor = 	 {III, Hal Daumé and Singh, Aarti},
  volume = 	 {119},
  series = 	 {Proceedings of Machine Learning Research},
  month = 	 {13--18 Jul},
  publisher =    {PMLR},
  url = 	 {https://proceedings.mlr.press/v119/cai20b.html}
}

@inproceedings{in_experiment,
author = {Deng, Alex and Du, Michelle and Matlin, Anna and Zhang, Qing},
title = {Variance Reduction Using In-Experiment Data: Efficient and Targeted Online Measurement for Sparse and Delayed Outcomes},
year = {2023},
isbn = {9798400701030},
publisher = {Association for Computing Machinery},
address = {New York, NY, USA},
url = {https://doi.org/10.1145/3580305.3599928},
doi = {10.1145/3580305.3599928},
booktitle = {Proceedings of the 29th ACM SIGKDD Conference on Knowledge Discovery and Data Mining},
pages = {3937–3946},
numpages = {10},
location = {Long Beach, CA, USA},
series = {KDD '23}
}

@inproceedings{cuped,
author = {Deng, Alex and Xu, Ya and Kohavi, Ron and Walker, Toby},
title = {Improving the sensitivity of online controlled experiments by utilizing pre-experiment data},
year = {2013},
isbn = {9781450318693},
publisher = {Association for Computing Machinery},
address = {New York, NY, USA},
url = {https://doi.org/10.1145/2433396.2433413},
doi = {10.1145/2433396.2433413},
booktitle = {Proceedings of the Sixth ACM International Conference on Web Search and Data Mining},
pages = {123–132},
numpages = {10},
location = {Rome, Italy},
series = {WSDM '13}
}

@book{ lattimore19bandit,
  author = "Tor Lattimore and Csaba Szepesvari",
  title = "Bandit Algorithms",
  publisher = "Cambridge University Press",
  year = "2019"
}

@inproceedings{sachdeva2020off,
  title={Off-policy bandits with deficient support},
  author={Sachdeva, Noveen and Su, Yi and Joachims, Thorsten},
  booktitle={Proceedings of the 26th ACM SIGKDD International Conference on Knowledge Discovery \& Data Mining},
  pages={965--975},
  year={2020}
}

@inproceedings{sakhi2023pac,
  title={{PAC-Bayesian Offline Contextual Bandits with Guarantees}},
  author={Sakhi, Otmane and Alquier, Pierre and Chopin, Nicolas},
  booktitle={International Conference on Machine Learning},
  pages={29777--29799},
  year={2023},
  organization={PMLR}
}

@inproceedings{sakhi2020blob,
  title={{BLOB: A Probabilistic model for recommendation that combines organic and bandit signals}},
  author={Sakhi, Otmane and Bonner, Stephen and Rohde, David and Vasile, Flavian},
  booktitle={Proceedings of the 26th ACM SIGKDD International Conference on Knowledge Discovery \& Data Mining},
  pages={783--793},
  year={2020}
}

@article{horvitz1952generalization,
  title={A generalization of sampling without replacement from a finite universe},
  author={Horvitz, Daniel G and Thompson, Donovan J},
  journal={Journal of the American statistical Association},
  volume={47},
  number={260},
  pages={663--685},
  year={1952},
  publisher={Taylor \& Francis}
}

@article{metelli2021subgaussian,
  title={Subgaussian and differentiable importance sampling for off-policy evaluation and learning},
  author={Metelli, Alberto Maria and Russo, Alessio and Restelli, Marcello},
  journal={Advances in Neural Information Processing Systems},
  volume={34},
  pages={8119--8132},
  year={2021}
}

@inproceedings{su2020doubly,
  title={Doubly robust off-policy evaluation with shrinkage},
  author={Su, Yi and Dimakopoulou, Maria and Krishnamurthy, Akshay and Dud{\'\i}k, Miroslav},
  booktitle={International Conference on Machine Learning},
  pages={9167--9176},
  year={2020},
  organization={PMLR}
}

@InProceedings{saito2022off,
  title = 	 {Off-Policy Evaluation for Large Action Spaces via Embeddings},
  author =       {Saito, Yuta and Joachims, Thorsten},
  booktitle = 	 {Proceedings of the 39th International Conference on Machine Learning},
  pages = 	 {19089--19122},
  year = 	 {2022},
  volume = 	 {162},
  series = 	 {Proceedings of Machine Learning Research},
  month = 	 {17--23 Jul},
  publisher =    {PMLR},
  url = 	 {https://proceedings.mlr.press/v162/saito22a.html}
}

@book{mdp,
author = {Puterman, Martin L.},
title = {Markov Decision Processes: Discrete Stochastic Dynamic Programming},
year = {1994},
isbn = {0471619779},
publisher = {John Wiley \& Sons, Inc.},
address = {USA},
edition = {1st}
}

@inproceedings{long_hor_cond,
author = {Liu, Yao and Bacon, Pierre-Luc and Brunskill, Emma},
title = {Understanding the curse of horizon in off-policy evaluation via conditional importance sampling},
year = {2020},
publisher = {JMLR.org},
booktitle = {Proceedings of the 37th International Conference on Machine Learning},
articleno = {574},
numpages = {10},
series = {ICML'20}
}

@inproceedings{curse_long,
 author = {Liu, Qiang and Li, Lihong and Tang, Ziyang and Zhou, Dengyong},
 booktitle = {Advances in Neural Information Processing Systems},
 editor = {S. Bengio and H. Wallach and H. Larochelle and K. Grauman and N. Cesa-Bianchi and R. Garnett},
 pages = {},
 publisher = {Curran Associates, Inc.},
 title = {Breaking the Curse of Horizon: Infinite-Horizon Off-Policy Estimation},
 volume = {31},
 year = {2018}
}

@InProceedings{cond_is,
  title = 	 {Conditional Importance Sampling for Off-Policy Learning},
  author =       {Rowland, Mark and Harutyunyan, Anna and van Hasselt, Hado and Borsa, Diana and Schaul, Tom and Munos, Remi and Dabney, Will},
  booktitle = 	 {Proceedings of the Twenty Third International Conference on Artificial Intelligence and Statistics},
  pages = 	 {45--55},
  year = 	 {2020},
  editor = 	 {Chiappa, Silvia and Calandra, Roberto},
  volume = 	 {108},
  series = 	 {Proceedings of Machine Learning Research},
  month = 	 {26--28 Aug},
  publisher =    {PMLR},
  url = 	 {https://proceedings.mlr.press/v108/rowland20b.html}
}

@inproceedings{gilotte2018offline,
  title={{Offline A/B testing for recommender systems}},
  author={Gilotte, Alexandre and Calauz{\`e}nes, Cl{\'e}ment and Nedelec, Thomas and Abraham, Alexandre and Doll{\'e}, Simon},
  booktitle={Proceedings of the Eleventh ACM International Conference on Web Search and Data Mining},
  pages={198--206},
  year={2018}
}

@InProceedings{aouali23a,
  title = 	 {{Exponential Smoothing for Off-Policy Learning}},
  author =       {Aouali, Imad and Brunel, Victor-Emmanuel and Rohde, David and Korba, Anna},
  booktitle = 	 {Proceedings of the 40th International Conference on Machine Learning},
  pages = 	 {984--1017},
  year = 	 {2023},
  publisher =    {PMLR},
}

\appendix

\section{Additional Discussions.}

\subsection{Our family of estimators is efficient}\label{app:conv_same_family}
Recall the definition of our family of estimators. Let $f$ be a bounded function. Our estimator is constructed using an $f$-regularized IPS estimator on $\mathcal{D}_B$, and a bias estimation using the trajectories of $\mathcal{D}_A$. If we reverse the roles of $\mathcal{D}_B$ and $\mathcal{D}_A$, we can construct with the help of another function $f'$, another estimator:
\begin{align*}
    \hat{\mathcal{I}}^R_{f'} &= \frac{1}{n_A} \sum_{i \in u_A}\sum_{t = 1}^{T_i} f'(w_B(\tau^i_t)) r^i_t \\ &+ \frac{1}{n_B} \sum_{j \in u_B} \sum_{t = 1}^{T_j} \left(w_A(\tau^j_t)\left[1 - f'(w_B(\tau^j_t))\right] - 1\right) r^j_t\,.
\end{align*}
This construction is also valid. It uses the $f'$-regularized IPS estimator on $\mathcal{D}_A$  and corrects its bias with interactions of $\mathcal{D}_B$. It seems equivalent to using $\hat{\mathcal{I}}_{f}$ with a specific function $f$. Let $z$ be the function defined as:
\begin{align*}
    \forall x \in \mathbb{R}^{+*},  z(x) = x(1 - f'(1/x)) - 1,
\end{align*}
we have for any $x \in \mathbb{R}^{+*}$:
\begin{align*}
      z(x) = x(1 - f'(1/x)) - 1 \iff f'(x) = 1 - x(1 + z(1/x)).
\end{align*}
One can show with simple manipulations that
\begin{align*}
    \hat{\mathcal{I}}_{z} = \hat{\mathcal{I}}^R_{f'}.
\end{align*}
This means that the reversed estimator $\hat{\mathcal{I}}^R_{f'}$ is equivalent to using a specific function $z$ for our estimator $\hat{\mathcal{I}}_{z}$. In addition, and as our estimator is linear in the function $f$, creating a convex combination between $\hat{\mathcal{I}}_{f}$ and $\hat{\mathcal{I}}^R_{f'}$ with the help of $\lambda \in [0, 1]$ is equivalent to the estimator $\hat{\mathcal{I}}_{\lambda f + (1 - \lambda)z}$ as we have for any $\lambda \in [0, 1]$:
\begin{align*}
    \lambda \hat{\mathcal{I}}_{f} + (1 - \lambda) \hat{\mathcal{I}}^R_{f'}&= \lambda \hat{\mathcal{I}}_{f} + (1 - \lambda) \hat{\mathcal{I}}_{z} \\
    &= \hat{\mathcal{I}}_{\lambda f + (1 - \lambda) z}.
\end{align*}
This proves that our parametrization is sufficient.

\subsection{Variance  and its surrogate} \label{app:variance_surrogate}
Our ultimate goal is to find the function $f$ that results in the estimator with the lowest variance. If we cannot find that, we will be happy about a function $f$ that drastically improves on the difference estimator. We already know that any function $f$, as long as it respects condition \eqref{eq:cond_2}, will do so when $\pi_A$ and $\pi_B$ are nearly identical. But that was also achieved by vanilla IPS and is not sufficient to build robust estimators. To simplify our study, we suppose that the covariances within trajectories can be neglected and are dominated by the sum of variances:
\begin{align*}
    \text{var}\left[\hat{\mathcal{I}}_{f}\right] 
    &\approx \frac{1}{n_B} \sum_{t = 1}^{\bar{T}} \text{var}_{\nu(\pi_B)}\left[f(w_A(\tau_t)) r_t \right] \\ &+ \frac{1}{n_A} \sum_{t = 1}^{\bar{T}}\text{var}_{\nu(\pi_A)}\left[\left(1 - w_B(\tau_t)\left[ 1 + f(w_A(\tau_t))\right] \right) r_t \right]\,.
\end{align*}
This holds when the states within trajectories present little correlations (a milder condition than contextual bandit setting \cite{lattimore19bandit}, or when the covariances nullify each other. With this assumption, we can upper bound the variance with our surrogate $S_f$ as we have:
\begin{align*}
\text{var}\left[\hat{\mathcal{I}}_{f}\right] 
    &\le S_f.
\end{align*}
The surrogate is an upper bound of the variance (when neglecting covariances within trajectories), and can be tight when:
\begin{align*}
    &\sum_{t = 1}^{\bar{T}}(\mathbb{E}_{\nu(\pi_A)}\left[\left(1 - w_B(\tau_t)\left[ 1 + f(w_A(\tau_t))\right] \right) r_t \right])^2 \\&+\sum_{t = 1}^{\bar{T}}(\mathbb{E}_{\nu(\pi_B)}\left[(f(w_A(\tau_t)) r_t)\right])^2 \ll S_f.
\end{align*}
This holds when the reward signal has a much smaller expectation that its variance. This is the case in computational advertising scenarios, where the CTR or conversion rates are small \cite{sakhi2020blob}. This term should also be negligible when $\pi_A$ and $\pi_B$ are close.

\subsection{Misspecification - Sensitivity Analysis}\label{app:misspecification}

We write $\sum_{B,t}$ for $\sum_{j\in u_B}\sum_{t=1}^{T_j}$ and $\sum_{A,t}$ similarly. The $f$-regularized estimator with misspecified weights is
\begin{align*}
\tilde{\mathcal I}_{f}
=
\frac{1}{n_B}\sum_{B,t} f(\hat w_A(\tau_t^j))r_t^j
+
\frac{1}{n_A}\sum_{A,t}
\left(
1-\hat w_B(\tau_t^i)
\left[1+f(\hat w_A(\tau_t^i))\right]
\right)r_t^i ,
\end{align*}
with $f(0)=-1$. Since the weights are misspecified, unbiasedness no longer holds. We quantify the induced bias in the small-misspecification regime. Let $\mathbb E_A=\mathbb E_{\nu(\pi_A)}$, $\mathbb E_B=\mathbb E_{\nu(\pi_B)}$, and let $D_t=\Delta(\tau_t)$. Under the multiplicative perturbation model, we have
\begin{align*}
&\hat w_A(\tau_t)=w_A(\tau_t)(1+\delta D_t)+O(\delta^2), \\
&\hat w_B(\tau_t)=w_B(\tau_t)(1-\delta D_t)+O(\delta^2).
\end{align*}
Assuming $f$ is differentiable, a first-order Taylor expansion gives
\[
f(\hat w_A(\tau_t))
=
f(w_A(\tau_t))
+
\delta w_A(\tau_t)f'(w_A(\tau_t))D_t
+
O(\delta^2).
\]
Therefore, denoting
$b_\delta(\Delta,f)=\mathbb E[\tilde{\mathcal I}_f(\hat w)]-\mathbb E[\tilde{\mathcal I}_f(w)]$, we obtain
\begin{align*}
b_\delta(\Delta,f)
=
\delta
\Bigg[
&
\mathbb E_B
\sum_{t=1}^T
w_A(\tau_t)f'(w_A(\tau_t))D_t r_t
\\
&+
\mathbb E_A
\sum_{t=1}^T
w_B(\tau_t)D_t
\Big(
1+f(w_A(\tau_t))
-w_A(\tau_t)f'(w_A(\tau_t))
\Big)r_t
\Bigg]
\\
&+
O(\delta^2).
\end{align*}
Using the change-of-measure identity
$\mathbb E_A[w_B(\tau_t)g(\tau_t)]=\mathbb E_B[g(\tau_t)]$,
the derivative terms cancel, yielding
\begin{align*}
b_\delta(\Delta,f)
=
\delta\,
\mathbb E_B
\left[
\sum_{t=1}^T
D_t
\left(1+f(w_A(\tau_t))\right)r_t
\right]
+
O(\delta^2).
\end{align*}
Thus, misspecification bias is controlled by making $f$ close to $-1$ on trajectories where the noise level $D_t$ is large.

Ignoring higher-order terms in $\delta$, the MSE decomposes as
\[
\mathrm{MSE}(\tilde{\mathcal I}_{f})
=
\mathrm{var}(\tilde{\mathcal I}_{f})
+
b_\delta(\Delta,f)^2 .
\]
The variance term is controlled by the surrogate $S_f$. It remains to control the squared bias. Let
\[
X_t
=
D_t\left(1+f(w_A(\tau_t))\right)r_t .
\]
From the first-order bias expansion,
\[
b_\delta(\Delta,f)
=
\delta\,\mathbb E_B\left[\sum_{t=1}^T X_t\right]
+
O(\delta^2).
\]
Dropping higher-order terms, Jensen's inequality gives
\[
b_\delta(\Delta,f)^2
\leq
\delta^2\,
\mathbb E_B\left[
\left(\sum_{t=1}^T X_t\right)^2
\right].
\]
Then, by Cauchy--Schwarz,
\[
\left(\sum_{t=1}^T X_t\right)^2
\leq
T\sum_{t=1}^T X_t^2 .
\]
Combining the two inequalities yields
\[
b_\delta(\Delta,f)^2
\leq
\delta^2 T\,
\mathbb E_B\left[
\sum_{t=1}^T
\left(
D_t(1+f(w_A(\tau_t)))r_t
\right)^2
\right]
=
\delta^2 T B_\Delta(f).
\]
Therefore,
\[
\mathrm{MSE}(\tilde{\mathcal I}_{f})
\lesssim
S_f+\delta^2T B_\Delta(f),
\]
where
\[
B_\Delta(f)
=
\mathbb E_B
\left[
\sum_{t=1}^T
\left(
D_t(1+f(w_A(\tau_t)))r_t
\right)^2
\right].
\]
This motivates the robust objective
\[
S_{\lambda,\Delta}(f)=S_f+\lambda T B_\Delta(f),
\]
where $\lambda\geq 0$ controls the bias--variance tradeoff. The objective separates over trajectory prefixes and is quadratic in each value $f(w_A(\tau_t))$. Setting the derivative to zero gives, with
$\gamma_t=\lambda T n_A D_t^2$,
\[
f^\star_{\lambda,\Delta}(w_A(\tau_t))
=
\frac{(1-\gamma_t)w_A(\tau_t)-1}
{(n_r+\gamma_t)w_A(\tau_t)+1}.
\]
Equivalently, one may absorb the factor $Tn_A$ into $\lambda$, in which case $\gamma_t=\lambda D_t^2$.

\subsection{Logarithmic noise and MLE}\label{app:noise_model}
A logarithmic noise model is natural when propensities are learned by maximum likelihood, because maximum likelihood controls errors on the log-probability scale. Let $\pi(a\mid s)$ denote a true propensity and $\hat\pi(a\mid s)$ its estimate, and define the log-propensity error
\[
\eta(a,s)=\log \hat\pi(a\mid s)-\log \pi(a\mid s).
\]
The excess negative log-likelihood is a KL divergence, which is locally quadratic in this log-error. Indeed, since $\hat\pi=\pi e^\eta$ and normalization imposes $\mathbb E_{\pi}[e^\eta]=1$, a second-order expansion gives
\[
\mathbb E_{\pi}\left[\log\frac{\pi(a\mid s)}{\hat\pi(a\mid s)}\right]
=
-\mathbb E_{\pi}[\eta(a,s)]
\approx
\frac12 \mathbb E_{\pi}[\eta(a,s)^2].
\]
Thus, locally, maximum likelihood controls squared log-propensity errors rather than squared errors in raw probabilities. This is aligned with importance weighting: for a one-step ratio $w=\pi_B/\pi_A$,
\[
\log \hat w-\log w
=
\left(\log \hat\pi_B-\log \pi_B\right)
-
\left(\log \hat\pi_A-\log \pi_A\right),
\]
and for a trajectory this error accumulates additively over time. In contrast, the corresponding error in the raw importance weight is multiplicative. This motivates modeling misspecification on the log-ratio scale.

The specific choice $\Delta(x)=\min(|\log x|,1)$ should therefore be viewed as a practical proxy rather than an exact model of the maximum-likelihood error. It uses the observed separation between the two policies on the log-ratio scale as a surrogate for where propensity-ratio errors are likely to matter. This proxy has three useful properties: it vanishes when the policies agree, grows smoothly as their propensities separate, and remains bounded because of clipping. It therefore preserves the variance gains in near-A/A regions while preventing extreme ratios from dominating the robustification.

\section{Detailed Experimental Setup and Results}\label{app:detailed_experiments}

% In this section, we provide more details about the experimental settings. We conduct experiments to empirically validate the behavior of our family of estimators and quantify the variance reduction $v(f) = \text{var}(\hat{\mathcal{I}})/\text{var}(\hat{\mathcal{I}_f})$ achieved by our family of estimators in different scenarios. We simulate an environment with a discrete action space $\mathcal{A}$ of size $\lvert\mathcal{A}\rvert = 10$ and binary, noisy sparse rewards reminiscent of online decision systems scenarios (e.g. Banner layout to increase CTR, or conversions). These rewards are Bernoulli variable of probabilities close to $p =10^{-2}$. The expectation of the rewards for each action $a$ changes from a setting to another, but stays in the $p =10^{-2}$ region.

\subsection{The bandit setting}\label{app:bandit_experiments}

% \begin{table}[t]
%     \centering

%     \caption{Bandit Setting: the CTRs $p(a)$ of the difference actions.}
%     \label{app_tab:ctr}
%     \begin{tabular}{c||ccccccccc}
%         \toprule
%          & $n_r$ & $v(h_1)$ &  $v(f^\star_1)$ &  $v(f^\star_{n_r})$ \\
%          $p$ & $a = 1$ & $a = 2$ &  $a = 3$ &  $v(f^\star_{n_r})$ \\
%         \bottomrule
%     \end{tabular}
    
%     \vspace{-0.5cm}
% \end{table}

\begin{figure}[ht]
    \centering
    \includegraphics[width=0.7\linewidth]{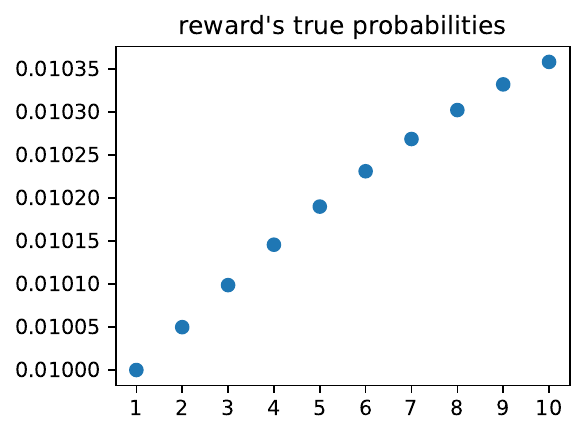}
    \caption{Rewards true probabilities $p$ for the bandit setting.}
    \label{fig:enter-label}
\end{figure}

We compute the distance $d(\pi_A, \pi_B)$ as:
\begin{align*}
    d(\pi_A, \pi_B) = \frac{1}{2} \left(\mathbb{E}_{a \sim \pi_A}\left[\left(\frac{\pi_B(a)}{\pi_A(a)} - 1\right)^2 \right]  + \mathbb{E}_{a \sim \pi_B}\left[\left(\frac{\pi_A(a)}{\pi_B(a)} - 1\right)^2 \right] \right)
\end{align*}

\subsubsection{Distribution of the different policies considered.}

\begin{enumerate}
    \item \textbf{First case.} $d(\pi_A, \pi_B) = 0.266$. The variance reduction in this setting is very substantial.
    \item \textbf{Second case.} $d(\pi_A, \pi_B) =  7.93$. The variance reduction here was still very significant.
    \item \textbf{Third case.} $d(\pi_A, \pi_B) =  79.43$. The variance reduction here is marginal.
\end{enumerate}

% \subsubsection{Data Imbalance}
% In this section, we want to show how data imbalance affect the variance of the estimators, confirming our intuition that the difference estimator suffers the variance of the smallest population, while the off-policy estimators, once similarities can be leveraged, will not suffer any additional variance as long as the total population size is constant. Table~\ref{tab_app:variance_bandit} summarizes these variances for the close tested policies and far tested policies. We can see that for the first, close policies setting, $\hat{\mathcal{I}}_{f^\star_{n_r}}$ does not suffer any additional variance, with it being constant no matter the imbalance in the data, making it adaptive while the classical difference estimator suffers the variance of the smallest population. Once the policies are very different, all estimators suffer in the same way, as our optimal estimator $\hat{\mathcal{I}}_{f^\star_{n_r}}$ cannot leverage similarities between datasets.

\begin{table}
    \centering
    \caption{Variances for different settings.}
    \label{tab_app:variance_bandit}
    \begin{tabular}{cc||cccc}
        \toprule
        $d(\pi_A,\pi_B)$ & $n_r$ & $\text{var}\left[\hat{\mathcal{I}} \right]$  &  $\text{var}\left[\hat{\mathcal{I}}_{h_1} \right]$  &  $\text{var}\left[\hat{\mathcal{I}}_{f^\star_1} \right]$ & $\text{var}\left[\hat{\mathcal{I}}_{f^\star_{n_r}} \right]$  \\
        \midrule
        \multirow{3}{*}{$0.266$} & $1/4$ & $0.0628$ & $0.0070$ & $0.0033$ & $0.0022$   \\
        
         & $1$ & $0.0402$ & $0.0030$ & $0.0021$ & $0.0021$ \\
        
         & $4$ & $0.0628$ & $0.0023$ & $0.0033$ & $0.0022$ \\

        \midrule
        
        \multirow{3}{*}{$79.10$} & $1/4$ & $0.0622$ & $0.0552$ & $0.0558$ & $0.0543$ \\
        
         & $1$ & $0.0400$ & $0.0361$ & $0.0352$ & $0.0352$  \\

         & $4$ & $0.0626$ & $0.0577$ & $0.0543$ & $0.0518$  \\
        \bottomrule
    \end{tabular}
    
\end{table}

\end{document}